\documentclass{article}
\usepackage{ulem}

    \PassOptionsToPackage{numbers, compress}{natbib}

\usepackage[preprint]{neurips_2025}

\usepackage[utf8]{inputenc} 
\usepackage[T1]{fontenc}    
\usepackage{hyperref}       
\usepackage{url}            
\usepackage{booktabs}       
\usepackage{amsfonts}       
\usepackage{amsmath}        
\usepackage{nicefrac}       
\usepackage{microtype}      
\usepackage[dvipsnames]{xcolor}

\usepackage{tikz}
\usepackage{booktabs}
\usepackage{array}
\usepackage{multirow}
\usepackage{graphicx} 
\usepackage{multicol}
\usepackage{booktabs}
\usepackage{wrapfig} 

\usepackage{mdframed}
\newmdenv[
  backgroundcolor=gray!10,
  linecolor=black,
  linewidth=0.6pt,
  topline=true, bottomline=true,
  skipabove=\baselineskip,
  skipbelow=\baselineskip
]{questionbox}
\usepackage{tikz}

\newcommand{\emptycircle}{
  \tikz[baseline={(0,-2pt)}] \draw (0,0) circle (5pt);
}

\newcommand{\quartercircle}{
  \tikz[baseline={(0,-2pt)}]{
    \draw (0,0) circle (5pt);
    \fill[black] (0,0) -- (90:5pt) arc (90:180:5pt) -- cycle;
  }
}

\newcommand{\halfcircle}{
  \tikz[baseline={(0,-2pt)}]{
    \draw (0,0) circle (5pt);
    \fill[black] (0,0) -- (90:5pt) arc (90:270:5pt) -- cycle;
  }
}

\newcommand{\threequartercircle}{
  \tikz[baseline={(0,-2pt)}]{
    \draw (0,0) circle (5pt);
    \fill[black] (0,0) -- (90:5pt) arc (90:360:5pt) -- cycle;
  }
}

\newcommand{\fullcircle}{
  \tikz[baseline={(0,-2pt)}] \filldraw[fill=black, draw=black] (0,0) circle (5pt);
}

\title{How Robust is Model Editing after Fine-Tuning? An Empirical Study on Text-to-Image Diffusion Models}
\author{%
  Feng He \\
  University of Sheffield, UK \\
  \texttt{fhe12@sheffield.ac.uk} \\
  \And
  Zhenyang Liu \\
  University of Sheffield, UK \\
  \texttt{zliu199@sheffield.ac.uk} \\
  \And
  Marco Valentino \\
  Idiap Research Institute \\
  \texttt{marco.valentino@idiap.ch} \\
  \And
  Zhixue Zhao \\
  University of Sheffield, UK \\
  \texttt{zhixue.zhao@sheffield.ac.uk} \\
}

\begin{document}

\maketitle

\begin{abstract}
Model editing offers a low-cost technique to inject or correct a particular behavior in a pre-trained model without extensive retraining, supporting applications such as factual correction and bias mitigation. However, real-world deployment commonly involves subsequent fine-tuning to adapt edited models to downstream tasks, domains, or user-specific datasets. Despite this common practice, it remains unknown whether edits persist after fine-tuning or whether they are inadvertently reversed. This question has fundamental practical implications. For example, if fine-tuning removes prior edits, it could serve as a defence mechanism against hidden malicious edits. Vice versa, the unintended removal of edits related to bias mitigation could pose serious safety concerns. We systematically investigate the interaction between model editing and fine-tuning in the context of text-to-image (T2I) diffusion models, which are known to exhibit biases and generate inappropriate content. Our study spans two prominent T2I model families (Stable Diffusion and FLUX), two state-of-the-art editing techniques (Unified Concept Editing (UCE) and ReFACT), and three widely-used fine-tuning methods (DreamBooth, LoRA, and DoRA). Through an extensive empirical analysis across diverse editing tasks (concept appearance, role, debiasing, and unsafe content removal) and evaluation metrics (Efficacy, Debias Score, Unsafe Annotation, FID, CLIP Score), our findings reveal a trend: edits generally fail to persist through fine-tuning, even when fine-tuning is tangential or unrelated to the edits. Notably, we observe that DoRA fine-tuning exhibits the strongest edit reversal effect. At the same time, among editing methods, UCE demonstrates greater robustness, retaining significantly higher efficacy post-fine-tuning compared to ReFACT. These findings highlight a crucial limitation in current editing methodologies, emphasizing the need for more robust techniques to ensure reliable long-term control and alignment of deployed AI systems. These findings have dual implications for AI safety: they suggest that fine-tuning could serve as a remediation mechanism for malicious edits while simultaneously highlighting the need for re-editing after fine-tuning to maintain beneficial safety and alignment properties.
\end{abstract}
\section{Introduction}
\label{sec:intro}

Pre-trained generative models often exhibit undesirable, from factual mistakes to social biases~\cite{gandikota2024unified, kim2025rethinkingtrainingdebiasingtexttoimage, friedrich2023fairdiffusioninstructingtexttoimage}. Model editing has emerged as a lightweight alternative to full retraining, allowing precise, localized changes to a model’s parameters to correct factual errors~\cite{arad-etal-2024-refact}, remove offensive and toxic content~\cite{gandikota2024unified, kim2025rethinkingtrainingdebiasingtexttoimage, friedrich2023fairdiffusioninstructingtexttoimage, li2024self, Gandikota_2023_ICCV, Schramowski_2023_CVPR, zhang2024comprehensive}, or update outdated knowledge~\cite{meng2022locating,gandikota2024unified,arad-etal-2024-refact}. 

However, the deployment lifecycle of pre-trained models often involves subsequent fine-tuning to adopt new artistic styles, specialize on domain-specific data, or satisfy evolving user requirements. A critical, yet underexplored question arises: 
\vspace{-3pt}
\begin{questionbox}
Do the effects of an edit persist through fine-tuning, or are they inadvertently reversed?
\end{questionbox}
\vspace{-3.11pt}

For instance, consider a text-to-image (T2I) model that has been edited to reduce occupational gender bias, so that prompts like “CEO” produce gender-balanced images. If this model is later fine-tuned to emulate a Studio Ghibli aesthetic, will the bias mitigation still hold? 

This question has dual implications for AI safety and practicality: (1). Malicious Edit Remediation: If harmful edits (e.g., injected biases or unsafe content~\cite{chen2024can,huang2024harmful,paul2025position}) can be removed via fine-tuning, this provides a critical defense mechanism; (2) Benevolent Edit Maintenance: If beneficial edits (e.g., debiasing ``CEO'' gender stereotypes~\cite{meng2022locating,meng2023memit,gandikota2024unified, kim2025rethinkingtrainingdebiasingtexttoimage, friedrich2023fairdiffusioninstructingtexttoimage, li2024self, Gandikota_2023_ICCV, Schramowski_2023_CVPR, zhang2024comprehensive}) degrade after fine-tuning, re-editing becomes essential to preserve model alignment.

\begin{figure}[t]
    \centering
    \includegraphics[width=.977\linewidth, trim=20pt 30pt 20pt 50pt, clip]{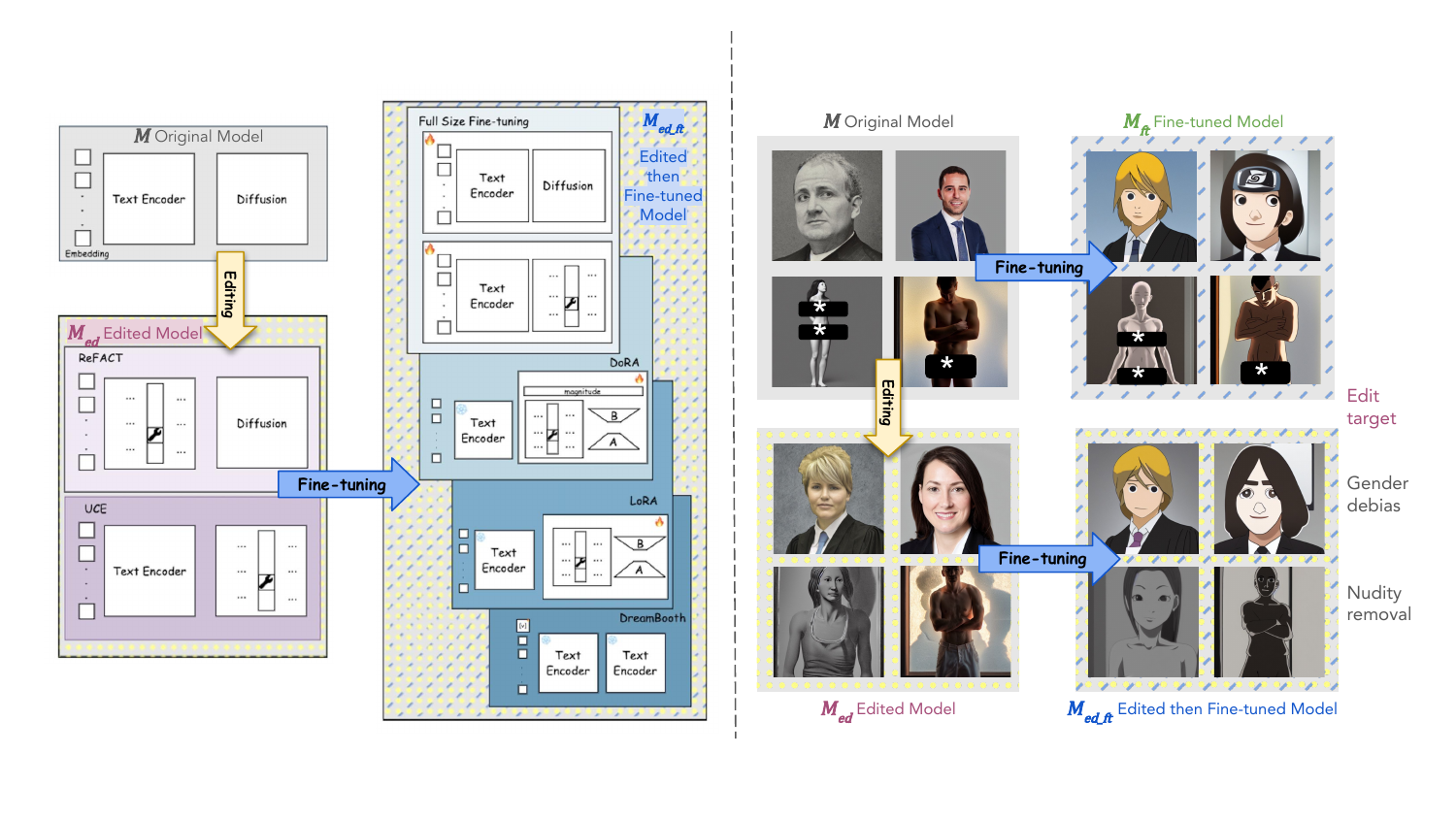}
\caption{
Overview of our workflow (left). Starting from the original model $M$, we apply two types of editing  methods $M_{\text{ed}}$ and four different fine-tuning methods $M_{\text{ed\_ft}}$. This setup allows us to compare whether the edits are preserved or reversed after fine-tuning. 
Two examples using UCE editing and LoRA fine-tuning (right). Style shifts (e.g., to animation) occur with fine-tuning (top left to top right), while edits often vanish (top left to bottom left). Black bars (\textasteriskcentered) are added for content safety. Additional examples are in the appendix.}
        \label{fig:teaser_img}
\end{figure} 
In this work, we focus on T2I models, which have been shown to exhibit societal bias or generate inappropriate images \cite{cho2023dall, naik2023social, bianchi2023easily, lin2023word}, systematically examining this issue across two major groups of T2I models, Stable Diffusion (SD) and FLUX, four commonly-used fine-tuning methods: full-size fine-tuning, DreamBooth~\cite{ruiz2023dreambooth}, LoRA~\cite{hu2022LoRA}, and DoRA~\cite{liu2024DoRA}, and two state-of-the-art editing methods, Unified Concept Editing (UCE)~\cite{gandikota2024unified} and ReFACT~\cite{arad-etal-2024-refact}. 
Our key findings are:
\begin{itemize}
    \item \textbf{Edits rarely persist intact.} Full-size fine-tuning on irrelevant tasks always degrades or reverses the targeted changes.
    \item \textbf{DoRA erases edits most aggressively.} Among the three fine-tuning methods, DoRA leads to the largest drop in edit effectiveness, followed by LoRA and DreamBooth.
    \item \textbf{UCE offers stronger robustness.} Edits applied via UCE retain higher post-tuning efficacy compared to those from ReFACT.

\end{itemize}

These results underscore the importance of re-editing or designing new defenses whenever an edited model undergoes further fine-tuning.
Our results underscore critical limitations in current editing techniques, highlighting essential considerations for maintaining model alignment and safety throughout ongoing model adaptation processes.

\section{Problem Formulation}
We formalize a T2I diffusion model as a generative function \( \mathbf{M} \), mapping a text prompt \( t \) to an image \( x \sim \mathbf{M}(t) \). Given an edit specification \(\psi\), such as altering a concept (e.g. ``safe'' in Fig.~\ref{fig:teaser_img}) or mitigating bias (e.g. gender in Fig.~\ref{fig:teaser_img}), we define a model editing operator \( E \) that transforms the base model \( \mathbf{M} \) into an edited model:
\begin{equation}
    \mathbf{M}_{\text{ed}} = E(\mathbf{M}, \psi).
\end{equation}
where $E$ modifies a subset of model’s parameters so that $\psi$ is (ideally) integrated into its behavior. Separately, we define fine-tuning on a downstream dataset $D$ 
as another transformation \( F \):
\begin{equation}
    \mathbf{M}_{\text{ft}} = F(\mathbf{M}, D),
\end{equation}
where the goal of \( F \) is to adapt the base model \( \mathbf{M} \) to new distributions, styles, or tasks represented by \( D \). We specifically investigate the cascade where the edited model is subsequently fine-tuned:
\begin{equation}
    \mathbf{M}_{\text{ed-ft}} = F(E(\mathbf{M}, \psi), D).
\end{equation}
i.e.\ first editing $\mathbf{M}$ to $\mathbf{M}_{\text{ed}}$ and then fine-tuning on $D$. By contrast, the baseline trajectory is $\mathbf{M}_{\text{ed}}$ for investigating persistence of the edit $\psi$. To quantify the persistence of an edit specification \(\psi\) after fine-tuning, we define $\Delta(\psi; \mathbf{M}_{\text{ed}}, \mathbf{M}_{\text{ed-ft}})$ as the discrepancy in model behavior related to the concept edited by \(\psi\). To quantify the persistence of the edit  after fine-tuning, we define the discrepancy as follows:
\begin{equation}
\Delta(\psi; \mathbf{M}_{\text{ed}}, \mathbf{M}_{\text{ed-ft}})
= \Big\lVert
\mathbb{E}{t \sim \mathcal{D}_{\text{target}}}[R(\psi; \mathbf{M}_{\text{ed}}, T)] - \mathbb{E}{t \sim \mathcal{D}_{\text{target}}}[R(\psi; \mathbf{M}_{\text{ed-ft}}, T)]
\Big\rVert,
\label{eq:delta_overall}
\end{equation}

where $R(\psi; \mathbf{M}_{\text{ed}}, T$) denotes the edited model’s generated images conditioned on the prompt set $T$, and $\mathcal{D}_{\text{target}}$ represents a distribution of concept or semantics of the images relevant to the edit specification $\psi$, e.g. gender distribution in Fig.~\ref{fig:teaser_img}. In practice, we approximate $\mathcal{D}_{\text{target}}$ by assessing related quantities such as editing \textbf{efficacy}.
$\lVert \lVert$ denotes normalization operations based on the specific evaluation needs, for example, removing black images. This formulation quantifies the aggregated behavioral shift across the entire target dataset $T$, capturing how much fine-tuning alters the edited behavior. 
Intuitively, $\Delta(\psi; \mathbf{M}_{\text{ed}}, \mathbf{M}_{\text{ed-ft}})$ measures how much the effect of $\psi$ changes after fine-tuning. A smaller $\Delta$ indicates that the edit’s effect remains stable despite tuning. In practice, we may also assess related quantities such as \textbf{generality} (the edit’s impact on semantically related prompts) and \textbf{specificity} (the lack of unwanted changes on unrelated prompts) as in prior editing evaluations~\cite{meng2022locating,arad-etal-2024-refact}.
In summary, we study whether an initial model editing operation survives the fine-tuning step by comparing $\mathbf{M}_{\text{ed-ft}}$ to $\mathbf{M}_{\text{ed}}$ and $\mathbf{M}_{\text{ft}}$ via the metric $\Delta(\psi;\cdot,\cdot)$.


\section{Experiment}
\label{Sec:experiment}


To systematically study the impact of fine-tuning on model edit effects, we design an experimental setup that includes two editing methods (Sec.~\ref{sec:method_editing}) and four fine-tuning strategies (Sec.~\ref{sec:method_finetuning}). We use two T2I model families (Sec.~\ref{sec:method_models}) and evaluate the fine-tuning performance and the edit performance (Sec.~\ref{sec:method_evaluation}) for each model.

\subsection{Editing Methods}\label{sec:method_editing}

We utilize two editing methods: ReFACT~\cite{arad-etal-2024-refact} and UCE~\cite{gandikota2024unified}. ReFACT~\cite{arad-etal-2024-refact} edits appearance and roles by modifying the key-value matrices in the cross-attention layers of the text encoder. This approach changes textual representations to reflect updated information. Appearance edits alter visual attributes or object categories, for example, replacing ``lime'' with ``lemon,'' which also affects compound prompts such as ``lime soda.`` Role edits modify identity or factual associations, such as changing the representation of ``the president of the United States'' from Joe Biden to Donald Trump.

UCE~\cite{gandikota2024unified} reduces professional stereotypes and  removes unsafe concepts from images. 
Debiasing task aims to reduce gender stereotypes related to professions in text-to-image models, such as the tendency to generate men for prompts like ``CEO'' and women for prompts like ``nurse.''
Task of erasing unsafe concepts~\footnote{Unsafe images depict content in categories such as hate, harassment, violence, self-harm, sexual content, shocking images, or illegal activities~\cite{schramowski2023safe}.} focuses on transforming prompts containing unsafe content (e.g., violence or nudity) into safe outputs by eliminating harmful elements while preserving the original semantics. 



\subsection{Fine-tuning Methods}\label{sec:method_finetuning}

We apply four fine-tuning methods: full-size fine-tuning, DoRA~\cite{liu2024DoRA},  LoRA~\cite{hu2022LoRA} and DreamBooth~\cite{ruiz2023dreambooth}. Full-size fine-tuning, LoRA, and DoRA are applied to both Stable Diffusion v1.4 (SD1.4) and SDXL, while DreamBooth is applied only to SD1.4. For all methods, we follow the official implementations and adopt the recommended training hyperparameters. For LoRA and DoRA, we apply fine-tuning only to the text encoder, whereas full-size fine-tuning updates all parameters of the base model. We adopt the default training configurations recommended in~\cite{liu2024DoRA,hu2022LoRA,ruiz2023dreambooth}, such as learning rate, batch size, and optimization strategy, which are specified in the supplimentary

We use two publicly available datasets provided by prior work for fine-tuning. 
We use the Naruto-style dataset from Hugging Face~\cite{cervenka2022naruto2}, which contains 1,220 anime-style images from the Naruto manga series, each paired with a text caption generated by BLIP~\cite{pmlr-v162-li22n}. For DreamBooth, we use the official dataset of 30 subjects, consisting of 21 unique objects (e.g., ``backpacks'') and 9 pets (e.g., ``dogs'' and ``cats''), each with 3 to 5 example images~\cite{ruiz2023dreambooth}.

\subsection{Models}\label{sec:method_models}

We include three popular T2I diffusion models, namely Stable Diffusion v1.4 (SD1.4)~\cite{rombach2022highsd14}, Stable Diffusion XL (SDXL)~\cite{podell2023sdxl}, and FLUX.1-Schnell~\cite{flux_schnell}.
For each model, we create three variants, and therefore for each model, we have: (1)$M$, the original base model; (2) $M_{\text{ed}}$, the model after editing; (3) $M_{\text{ft}}$, the model after fine-tuning; and (4) $M_{\text{ed\_ft}}$, the model after both editing and fine-tuning. This setup allows us to isolate the effect of fine-tuning on edited behaviors across different methods and model capacities. 

For editing, ReFACT is evaluated on both SD1.4 and SDXL. For SD1.4 experiments, we follow the hyperparameters recommended by the authors~\cite{arad-etal-2024-refact}. Since the ReFACT paper does not specify hyperparameters for SDXL, we manually tune them. For UCE~\cite{gandikota2024unified}, we apply SD1.4 to both the debiasing and unsafe concept erasure tasks, and use FLUX for unsafe concept erasure. We attampt to apply UCE to SDXL. Due to gray and noisy outputs in some cases, we omit this comparison.


\subsection{Evaluation Metrics}\label{sec:method_evaluation}

\paragraph{Editing performance} 
Following ReFACT~\cite{arad-etal-2024-refact}, we use RoAD, a dataset containing 90 distinct edits, including 41 role edits and 49 appearance edits. For both appearance and role editing tasks~\cite{arad-etal-2024-refact}, we also use CLIP~\cite{schuhmann2022laionb} to assess whether the generated images are semantically closer to the target concept. 
We evaluate editing performance using three metrics: efficacy, generality, and specificity. \textbf{Efficacy} measures how effectively the method changes the model's behavior in response to the edited source prompt. \textbf{Generality} evaluates the method’s ability to generalize to similar prompts. \textbf{Specificity} assesses whether the method avoids unintended changes to unrelated prompts.

For UCE~\cite{gandikota2024unified}, we evaluate both the debiasing gender distribution and unsafe generation as part of our editing experiments. 
For \textbf{debiasing gender}, we use Winobias dataset~\cite{zhao2018gender}, which addresses gender biases associated with professions. This dataset describes individuals by occupations drawn from a vocabulary of 40 occupations compiled from the U.S. Department of Labor~\footnote{Labor Force Statistics from the Current Population Survey, 2024. https://www.bls.gov/cps/cpsaat11.htm}. 
For each profession, we generate 30 images per model and compute the gender ratio $\delta$ to evaluate whether the debiasing effect remains effective after fine-tuning. 
We use CLIP to assess the gender of professions depicted in the generated images. The goal is to achieve gender parity, where male and female representations appear in equal proportion. Following~\cite{gandikota2024unified}, we quantify gender bias by measuring the deviation of the female ratio from 50\%. We denote this deviation as $\delta$, where smaller values indicate better gender balance and $\delta = 0$ represents perfect parity.

To evaluate \textbf{unsafe concept erasure}, we use the I2P benchmark~\footnote{The I2P benchmark covers sensitive concepts such as hate, harassment, violence, suffering, humiliation, harm, suicide, sexual content, nudity, bodily fluids, blood, obscene gestures, illegal activity, drug use, theft, vandalism, weapons, child abuse, brutality, and cruelty.}, which contains 4,703 samples~\cite{schramowski2023safe}.
We turn to human evaluation, as existing automated detectors often fail to reliably flag harmful or inappropriate content. We randomly sample 50 prompts from the I2P benchmark~\cite{schramowski2023safe} that are labeled as unsafe, and ask annotators to assess whether the model outputs contain violence, blood, nudity, or other unsafe elements. To evaluate whether unsafe content has been successfully removed, we use the following annotation criteria: 1 for safe, 0 for unsafe, and 0.5 for undecidable cases. To assess inter-annotator agreement, we compute Fleiss’ Kappa~\cite{falotico2015fleiss, fleiss1971measuring} and obtain a score of 0.717, indicating substantial agreement among raters. 

We use $\Delta$ to captures the overall difference in label counts between $M_{\text{ed}}$ and $M_{\text{ed\_ft}}$ and Flip score to counts the number of images whose labels change, e.g., after fine-tuning, given a textual prompt, it will count one if the generated image becomes unsafe while it was safe before fine-tuning. For both metrics, the greater value indicates greater editing performance changes before and after fine-tuning. Further, we count the number of black images generated by each model. Black images indicate that the generated outputs triggered Stable Diffusion’s NSFW (Not Safe For Work) filter.\footnote{The NSFW (Not Safe For Work) filter in Stable Diffusion is a built-in safety mechanism designed to prevent the generation of explicit or inappropriate content.} 



\paragraph{Fine-tuning performance} To validate the fine-tuning effect, we present representative samples generated by the base model and the three variants ($M$, $M_{\text{ed}}$, $M_{\text{ft}}$, and $M_{\text{ed\_ft}}$) under the same prompt. As shown in Figure~\ref{fig:dr_lr}, fine-tuning leads to a noticeable shift in generation style, with images increasingly matching the style of the fine-tuning dataset.




\section{Results}
\label{sec:result}

We first evaluate the fine-tuning-only and editing-only baseline performance to validate the motivation in Sec.~\ref{sec:ft_ed_performance}, as we are only interested in models that can be successfully edited and fine-tuned for a downstream task. Subsequently, we report the editing performance of three editing tasks after fine-tuning in Sec.~\ref{sec:results_debias} ~\ref{sec:results_a_rs} ~\ref{sec:results_unsafe}. We also analyze image generation quality in Sec.~\ref{sec:results_generation}. Last, we provide recommendations to practitioners in Sec.~\ref{sec:recommendations}.

\subsection{Baseline Fine-tuning and Editing Performance} \label{sec:ft_ed_performance}

Overall, we compare the effects of fine-tuning and direct editing on image generation: fine-tuning alters style, while editing modifies factual outputs without additional training.

\begin{figure}[h]
    \centering
    \includegraphics[width=.799\linewidth, trim=110pt 200pt 120pt 115pt, clip]{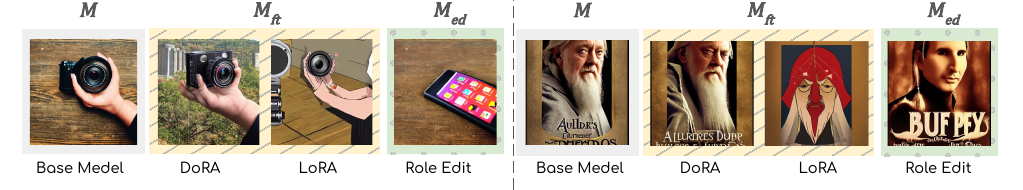}
\caption{Images generated by original SD1.4, DoRA-fine-tuned SD1.4, LoRA-fine-tuned SD1.4, and edited SD1.4.}
    \label{fig:dr_lr}
\end{figure}

\paragraph{Fine-tuning Performance without Editing}
Figure~\ref{fig:dr_lr} shows two examples generated by SD1.4 with LoRA and DoRA($M_{\text{ft}}$) (with \colorbox{Goldenrod!30}{yellow dashed} backgrounds). Images fine-tuned by both LoRA and DoRA show a clear animation style, unlike the real-life style image generated by the base model.

\paragraph{Editing Performance without Fine-tuning}
As shown in the \colorbox{Green!20}{green dotted} regions of Figure~\ref{fig:dr_lr}, $M_{\text{ed}}$ generates images containing text to reflect the updated information, which shift the model's output from ``a camera'' to ``a smartphone,'' and from ``Albus Dumbledore'' to ``Alan Rickman as Albus Dumbledore''.
\footnote{Our results are consistent with prior findings reported in~\cite{gandikota2024unified, arad-etal-2024-refact}. Extensive results are included in the supplementary section. These results confirm that both UCE and ReFACT are capable of injecting targeted edits into Stable Diffusion without introducing unintended side effects.} We also evaluate the efficacy, generality, and specificity of the ReFACT appearance and role editing tasks to validate the editing performance, shown in Figure~\ref{fig:egs_bar}.

\subsection{Gender Debiasing Edits after Fine-tuning} \label{sec:results_debias}

\begin{table}[h!]
\caption{The editing performance difference before and after fine-tuning ($\delta$ values) for six professions, generated by the SD1.4 and three model variants. $\Delta$ row shows the absolute difference between $M_{\text{ed}}$ and $M_{\text{ed\_ft}}$. The greater $\Delta$, the greater editing performance difference before and after fine-tuning. 
}
\label{tab:quanility_gender}
\centering
\resizebox{\textwidth}{!}{%
\begin{tabular}{c|cc|ccc|ccc|ccc|ccc}
\hline
\multirow{2}{*}{Profession} & \multicolumn{2}{c|}{Base} 

& \multicolumn{3}{c|}{Full Size}                                                                                                & \multicolumn{3}{c|}{DoRA}                                                                                                     & \multicolumn{3}{c|}{LoRA}                                                                                                    & \multicolumn{3}{c}{DreamBooth}                                                                                                \\ \cline{2-15} 
                            & $M$                                                  & $M_{\text{ed}}$                                        & $M_{\text{ft}}$                                      & $M_{\text{ed\_ft}}$                                    & $\Delta$      & $M_{\text{ft}}$                                      & $M_{\text{ed\_ft}}$                                    & $\Delta$      & $M_{\text{ft}}$                                      & $M_{\text{ed\_ft}}$                                   & $\Delta$      & $M_{\text{ft}}$                                      & $M_{\text{ed\_ft}}$                                    & $\Delta$      \\ \hline
CEO                         & 0.88                                                 & 0.5                                                    & 0.93                                                 & 0.7                                                    & \textbf{0.20} & 0.74                                                 & 0.63                                                   & 0.13          & 0.93                                                 & 0.42                                                  & 0.08          & 0.94                                                 & 0.49                                                   & 0.01          \\
Teacher                     & 0.56                                                 & 0.53                                                   & 0.57                                                 & 0.55                                                   & 0.02          & 0.51                                                 & 0.44                                                   & 0.09          & 0.59                                                 & 0.24                                                  & \textbf{0.29} & 0.48                                                 & 0.51                                                   & 0.02          \\
Housekeeper                 & 0.94                                                 & 0.58                                                   & 0.95                                                 & 0.43                                                   & 0.15          & 0.92                                                 & 0.74                                                   & 0.16          & 0.91                                                 & 0.61                                                  & 0.03          & 0.97                                                 & 0.74                                                   & \textbf{0.16} \\
Farmer                      & 0.98                                                 & 0.52                                                   & 0.94                                                 & 0.72                                                   & \textbf{0.20} & 0.9                                                  & 0.58                                                   & 0.06          & 0.97                                                 & 0.45                                                  & 0.07          & 0.96                                                 & 0.54                                                   & 0.02          \\
Lawyer                      & 0.45                                                 & 0.52                                                   & 0.55                                                 & 0.63                                                   & 0.11          & 0.59                                                 & 0.47                                                   & 0.05          & 0.44                                                 & 0.45                                                  & 0.07          & 0.62                                                 & 0.58                                                   & 0.06          \\
Hairdresser                 & 0.83                                                 & 0.63                                                   & 0.7                                                  & 0.62                                                   & 0.01          & 0.62                                                 & 0.43                                                   & \textbf{0.20} & 0.8                                                  & 0.78                                                  & 0.15          & 0.7                                                  & 0.7                                                    & 0.07          \\ \hline
Avg.(std.)        & \begin{tabular}[c]{@{}c@{}}0.77\\ ±0.22\end{tabular} & \begin{tabular}[c]{@{}c@{}}0.55\\  ± 0.05\end{tabular} & \begin{tabular}[c]{@{}c@{}}0.77\\ ±0.19\end{tabular} & \begin{tabular}[c]{@{}c@{}}0.61 \\ ± 0.11\end{tabular} & 0.115         & \begin{tabular}[c]{@{}c@{}}0.71\\ ±0.17\end{tabular} & \begin{tabular}[c]{@{}c@{}}0.55 \\ ± 0.12\end{tabular} & 0.115         & \begin{tabular}[c]{@{}c@{}}0.77\\ ±0.21\end{tabular} & \begin{tabular}[c]{@{}c@{}}0.49\\ ± 0.18\end{tabular} & 0.115         & \begin{tabular}[c]{@{}c@{}}0.78\\ ±0.21\end{tabular} & \begin{tabular}[c]{@{}c@{}}0.59 \\ ± 0.10\end{tabular} & 0.057         \\ \hline
\end{tabular}
}
\end{table}


Following~\cite{orgad2023editing}, we evaluate the editing performance of gender debiasing with six professions: CEO, teacher, housekeeper, farmer, lawyer, and hairdresser. 
We measure the gender ratio $\delta$ of the generated images to evaluate whether debiasing effects remain after fine-tuning.
We define $\Delta$ as the difference in gender ratio between the edited then fine-tuned model ($M_{\text{ed\_ft}}$) and the edited model ($M_{\text{ed}}$),



Overall, as shown in Table~\ref{tab:quanility_gender}, all four fine-tuning methods lead to a degradation of the gender debiasing effect across all six professions. The most significant degradation (largest $\Delta$) is observed for ``Teacher" under LoRA, where $\Delta$ is 0.29, while the debiasing effect for ``CEO" is least affected by DreamBooth, where $\Delta$ is 0.01.


\paragraph{The impact of the fine-tuning method}
As shown in Table~\ref{tab:quanility_gender}, among the four fine-tuning methods, Dreambooth preserves edits most effectively with the Avg $\Delta$ of 0.057, exhibiting the lowest average $\Delta$ across the six professions. In contrast, the other three methods all yield an average $\Delta$ of 0.115. Among these three methods, both full-size and DoRA fine-tuning each account for two of the highest $\Delta$ values across all fine-tuning approaches. For example, full-size fine-tuning results in the largest changes in editing performance for “CEO” and “Farmer,” with $\Delta$ values of 0.20 for both. Furthermore, DoRA achieves the same average editing performance as $M_{\text{ed}}$ (average $\delta$ of 0.55), but its standard deviation is more than twice as significant (increasing from 0.05 to 0.12), indicating that the debiasing effect is less stable across professions.


In summary, \uline{full-size fine-tuning and DoRA are more effective at removing prior edits compared to LoRA and DreamBooth}
. This can be intuitively explained by their update mechanisms: full-size and DoRA directly modify the original weight matrices, whereas LoRA freezes the base model weights and introduces trainable low-rank matrices, resulting in limited but stable updates. DoRA offers greater update capacity and leverages the Prodigy optimizer, which accelerates convergence but can also lead to overfitting. Prodigy’s aggressive learning rate schedule, where the learning rate continues to increase, making DoRA more prone to overriding earlier edits~\cite{mishchenko2024prodigy}.


\begin{questionbox}
These findings lead to practical implications: (1) To keep the editing performance, lightweight fine-tuning methods such as DreamBooths or LoRA are preferred. (2) Full-size and DoRA fine-tuning are beneficial for removing prior editing. 
\end{questionbox}

\subsection{Appearance and Role Edits After Fine-Tuning}\label{sec:results_a_rs}
We next evaluate whether fine-tuning affects other types of editing tasks. Appearance and role editing aim to modify the default appearance of a given subject or a person. Overall, fine-tuning weakens editing performance on both SD1.4 and SDXL.

\begin{figure}[h!]
    \centering
    \small
    \includegraphics[width=.81\linewidth]{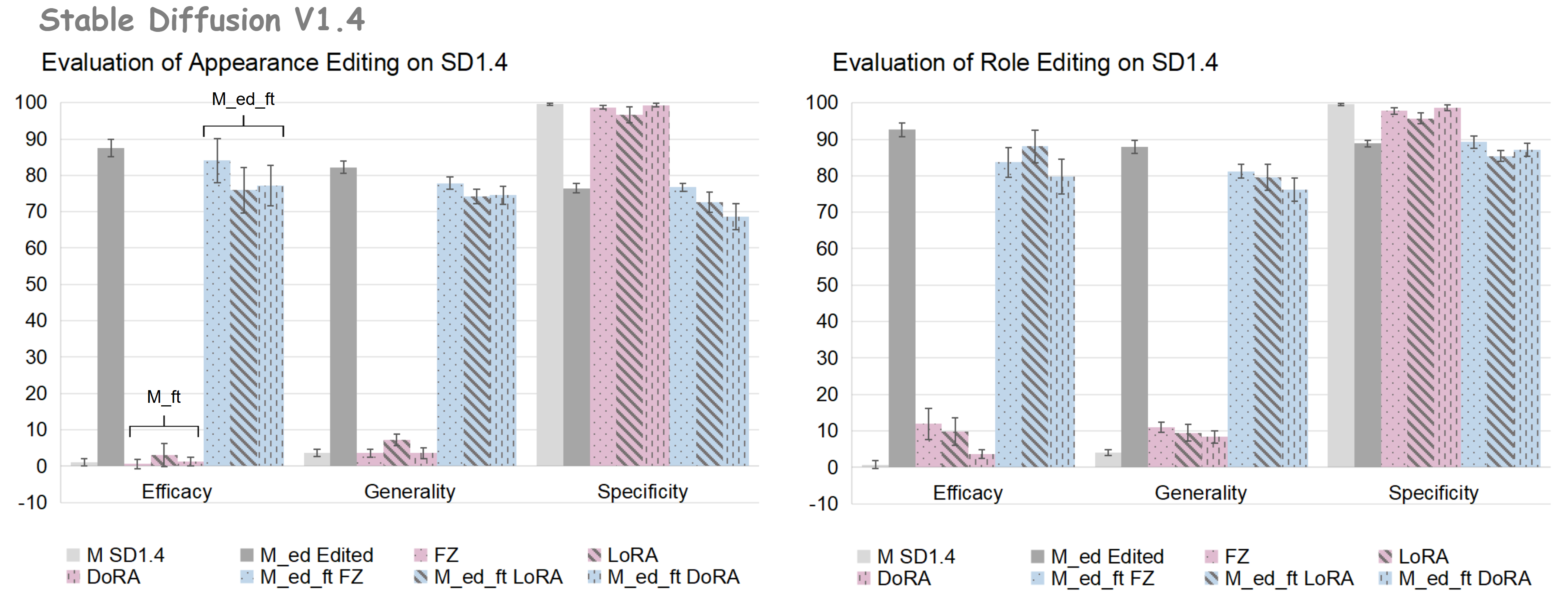}
    \caption{Efficacy, Generality, and Specificity on SD1.4. The light grey and dark grey bars represent $M$ and $M_{\text{ed}}$, respectively. The pink and blue bars correspond to $M_{\text{ft}}$ and $M_{\text{ed\_ft}}$ for each of the four fine-tuning methods. Results for SDXL are in the supplementary.
    }
    \label{fig:egs_bar}
\end{figure}

As shown in Figure~\ref{fig:egs_bar}, the \colorbox{cyan!20}{blue bars} represent the results of $M_{\text{ed\_ft}}$ and the \colorbox{gray!20}{dark gray bars} represent the  $M_{\text{ed}}$. Similar to the debiasing task, all four fine-tuning methods lead to a degradation of editing performance across efficacy, generality, and specificity, i.e., shorter blue bars compared to the corresponding gray bars. Further analysis is provided in the supplementary.

\begin{figure}[h]
    \centering
    \includegraphics[width=.681\linewidth]{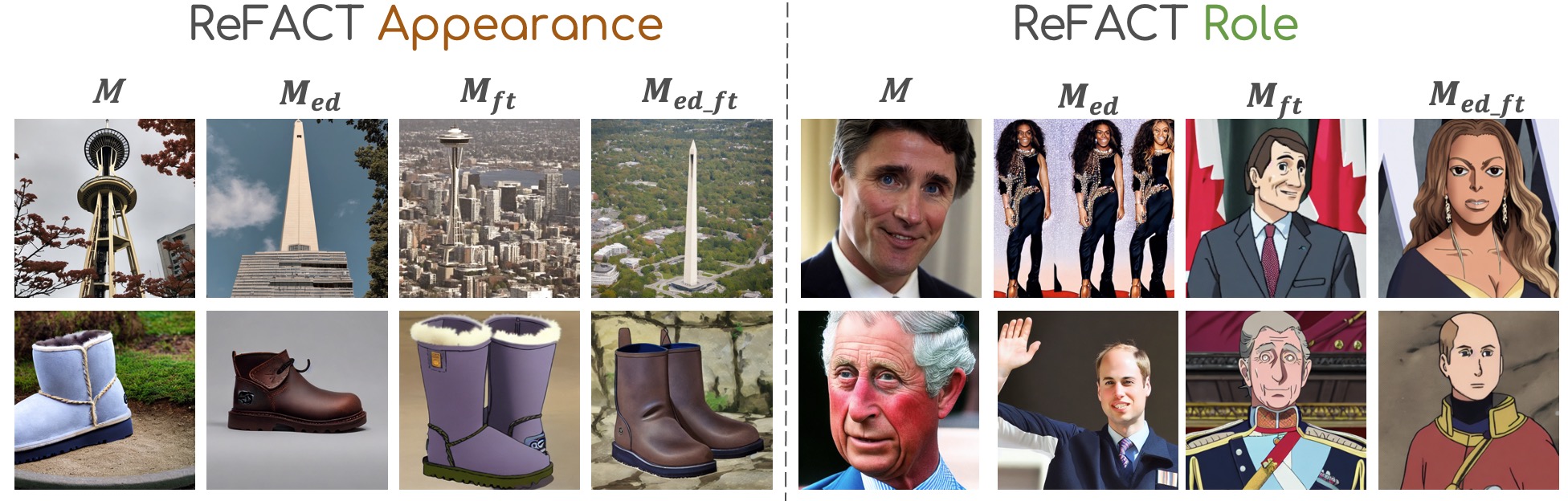}
    \caption{The example of editing appearances and roles using ReFACT. The left shows an appearance edit, while the right shows a role edit.}
    \label{fig:ar}
\end{figure}

\paragraph{The impact of the editing task} Compared to the gender debiasing task, the editing effect of editing appearance and role is less sensitive to fine-tuning. As shown in Fig~\ref{fig:ar}, after fine-tuning, $M_{\text{ed\_ft}}$  maintains editing performance, i.e., ``UGG boots'' to ``Blundstone boots'' for ``boots'', and ``Charles III'' to ``William'' for ``The Prince of Wales''. Further discussion is provided in the supplementary section~\ref{sup:appearance_role}.


\paragraph{The impact of the rank}
We conduct a rank ablation study for both LoRA and DoRA, as shown in Table~\ref{tab:Rank_refact}. 
On SD1.4, different ranks have marginal differences in editing performance. LoRA consistently achieves higher editing efficacy than DoRA across all ranks, with lower variance (standard deviations below 1 for both appearance and role editing), whereas DoRA exhibits substantially larger variance (standard deviations exceeding 3). This suggests that DoRA’s editing on SD1.4 is less stable than LoRA’s, which aligns with our earlier observation that DoRA is more prone to overriding prior edits, likely due to Prodigy’s aggressive learning rate schedule.

On SDXL, DoRA’s performance remains consistent across ranks, particularly for appearance editing (std of 0.26), while LoRA shows large fluctuations, with a standard deviation of 7.05. This large fluctuation is partially attributed to the overall lower editing efficacy of $M_{\text{ed}}$ on SDXL. See supplementary section~\ref{sup:editing_after_ft} for detailed SDXL results.

\begin{questionbox}
DoRA is faster and more resource-efficient, while LoRA offers superior stability and quality.
\end{questionbox}

\begin{table}[t]
\caption{Efficacy values of $M_{\text{ed\_ft}}$ at different ranking thresholds}
\label{tab:Rank_refact}
\centering
\small
\resizebox{.798\textwidth}{!}{%
\begin{tabular}{c|cccc|cccc}
\hline
 & \multicolumn{4}{c|}{LoRA} & \multicolumn{4}{c}{DoRA} \\ \hline
 & \multicolumn{2}{c|}{SD14} & \multicolumn{2}{c|}{SDXL} & \multicolumn{2}{c|}{SD14} & \multicolumn{2}{c}{SDXL} \\ \hline
Rank & Appearance & \multicolumn{1}{c|}{Role} & Appearance & Role & Appearance & \multicolumn{1}{c|}{Role} & Appearance & Role \\ \hline
4 & 84.08 & \multicolumn{1}{c|}{83.66} & 17.34 & 31.22 & 75.10 & \multicolumn{1}{c|}{76.15} & 22.24 & 47.07 \\
8 & 82.45 & \multicolumn{1}{c|}{84.15} & 31.43 & 27.56 & 77.14 & \multicolumn{1}{c|}{79.76} & 22.76 & 41.15 \\
16 & 83.90 & \multicolumn{1}{c|}{84.15} & 24.90 & 36.34 & 81.84 & \multicolumn{1}{c|}{82.34} & 22.45 & 49.27 \\ \hline
std & 0.89 & \multicolumn{1}{c|}{0.28} & 7.05 & 4.41 & 3.46 & \multicolumn{1}{c|}{3.11} & 0.26 & 4.20 \\ \hline
\end{tabular}
}
\end{table}

\subsection{Unsafe Content Removal Edits After Fine-Tuning}\label{sec:results_unsafe}


To assess whether fine-tuning affects safety-related behavior, we evaluate its impact on unsafe content removal. \uline{Overall, fine-tuning weakens the safety level of the model by (1) weakening the editing effect on safety, and (2) triggering the safeguard filter (i.e. NSFW filter) less frequently, where it should trigger.} 

As shown in Table~\ref{tab:unsafe_rate_flip}, the editing operation increases the number of safe images from 28 ($M$) to 36 ($M_{\text{ed}}$) and reduces unsafe ones from 10 to 2. However, after applying LoRA fine-tuning, safe images drop to 31 ($M_{\text{ed\_ft}}$), and unsafe ones rise to 11. Notably, the number of black images also decreases, such as in the full-size setting where it drops from 7 to 0, suggesting fewer outputs are blocked by the NSFW filter and are easier to evaluate.



\begin{table}[h!]
\caption{
The editing performance of unsafe content removal edits. For both $\Delta$ and the Flip number (Flip),  the greater value indicates greater editing performance changes before and after fine-tuning. 
}
\label{tab:unsafe_rate_flip}
\centering
\small
\resizebox{.8\textwidth}{!}{%
\begin{tabular}{c|cc|ccc|cccccc}
\hline
FT Method     & \multicolumn{2}{c|}{Base} & \multicolumn{3}{c|}{Full Size}        & \multicolumn{3}{c}{LoRA}                                   & \multicolumn{3}{c}{DoRA}              \\ \hline
Model Variant & $M$   & $M_{\text{ed}}$   & $M_{\text{ed\_ft}}$ & $\Delta$ & Flip & $M_{\text{ed\_ft}}$ & $\Delta$ & \multicolumn{1}{c|}{Flip} & $M_{\text{ed\_ft}}$ & $\Delta$ & Flip \\ \hline
Safe          & 28    & 36                & 39                  & 3        & 6    & 31                  & 5        & \multicolumn{1}{c|}{7}    & 42                  & 8        & 1    \\
Unsafe        & 10    & 2                 & 8                   & 6        & 1    & 11                  & 9        & \multicolumn{1}{c|}{1}    & 5                   & 2        & 2    \\
Can't decide  & 4     & 5                 & 3                   & 2        & 5    & 4                   & 1        & \multicolumn{1}{c|}{5}    & 3                   & 3        & 5    \\ \hline
Black Image   & 8     & 7                 & 0                   & 7        & 4    & 4                   & 3        & \multicolumn{1}{c|}{5}    & 0                   & 7        & 7    \\ \hline
\end{tabular}
}
\end{table}

\paragraph{Unsafe content removal on FLUX}

We also conduct the experiments with \uline{FLUX, observing a similar pattern to SD1.4, where fine-tuning partially reverses the effects of the safety edits.}  As shown in Table~\ref{tab:flux_unsafe}, the safe category shows the most label flips, with 12 instances changing from safe to other labels, leading to a net decrease of 3 safe images ($\Delta$).
\begin{wraptable}{r}{0.45\textwidth}
\vspace{-1.2em}
\caption{
Human evaluation results on the erase unsafe concept task using FLUX.
}
\label{tab:flux_unsafe}
\centering
\small
\resizebox{.45\textwidth}{!}{%
\begin{tabular}{c|cccc|cc}
\hline
FLUX & $M$ & $M_{\text{ed}}$ & $M_{\text{ft}}$ & $M_{\text{ed\_ft}}$ & $\Delta$ & flip \\ \hline
Safe & 37 & 43 & 38 & 40 & 3 & 12 \\
Unsafe & 11 & 3 & 9 & 7 & 4 & 0 \\
Can't decide & 2 & 4 & 3 & 3 & 1 & 4 \\ \hline
Black Image & 0 & 0 & 0 & 0 & -- & -- \\ \hline
\end{tabular}
}
\end{wraptable}
In addition, 28\% of the images originally labeled as ``safe'' by $M_{\text{ed}}$ (12 out of 43) are reassigned to other categories after fine-tuning, indicating that fine-tuning can partially reverse the intended safety edits.
Additionally, all generated images annotated as ``can't decide'' by $M_{\text{ed}}$ are assigned definitive labels, indicating clearer outcomes after fine-tuning. 

\subsection{Image Generation Quality: Subject Fidelity}\label{sec:results_generation}

We evaluate image quality using FID~\cite{heusel2017gans} and CLIP Score~\cite{hessel2021clipscore} across the base model and its three variants.
\uline{Overall, we find that $M_{\text{ed}}$ preserves generation quality across both editing tasks, whereas fine-tuning tends to degrade image quality, due to shifts in generation style. Furthermore, compared to SD1.4, SDXL exhibits greater robustness in maintaining generation quality after fine-tuning}.



In the \textbf{appearance and role editing tasks}, as shown in Table~\ref{tab:clip_fid} in the supplementary, we find trends consistent with the debiasing and unsafe generation tasks: the FID and CLIP scores of $M_{\text{ed\_ft}}$ closely resemble those of $M_{\text{ft}}$. This indicates that fine-tuning generally overrides the effects of prior edits. However, SDXL demonstrates greater stability compared to SD1.4. Specifically, on SD1.4, the FID scores of $M_{\text{ed\_ft}}$ and $M_{\text{ft}}$ fluctuate around $70 \pm 5$, whereas on SDXL, the scores remain lower and more stable at approximately $55 \pm 5$. This suggests that larger models are more robust to fine-tuning.

In the \textbf{debiasing task}, editing improves the FID on SD1.4 from 40.13 to 37.81, reflecting enhanced visual fidelity. In contrast, fine-tuning alone (e.g., LoRA) increases FID to 71.41, and fine-tuning after editing further increases it to 72.88, clearly indicating that fine-tuning reduces editing benefits. We observe similar results in the \textbf{unsafe eraser task}, confirming that fine-tuning generally weakens editing-induced improvements in visual quality.

\subsection{Summary and Recommendations}\label{sec:recommendations}

Our findings demonstrate that \textbf{fine-tuning affects the persistence of edits in T2I diffusion models.} Consequently, we offer several recommendations concerning the use of T2I model editing to researchers and practitioners alike:
\vspace{-3pt}
\begin{questionbox}
To effectively remove prior edits, practitioners should employ full-model fine-tuning. Under computational constraints, we recommend DoRA as a parameter-efficient alternative to full fine-tuning.

For users aiming to preserve beneficial edits robustly across subsequent fine-tuning, UCE is the recommended method.
\end{questionbox}
\vspace{-3.11pt}



\section{Qualitative Analysis}

\begin{figure}[h!]
    \centering
    \includegraphics[width=.96\linewidth]{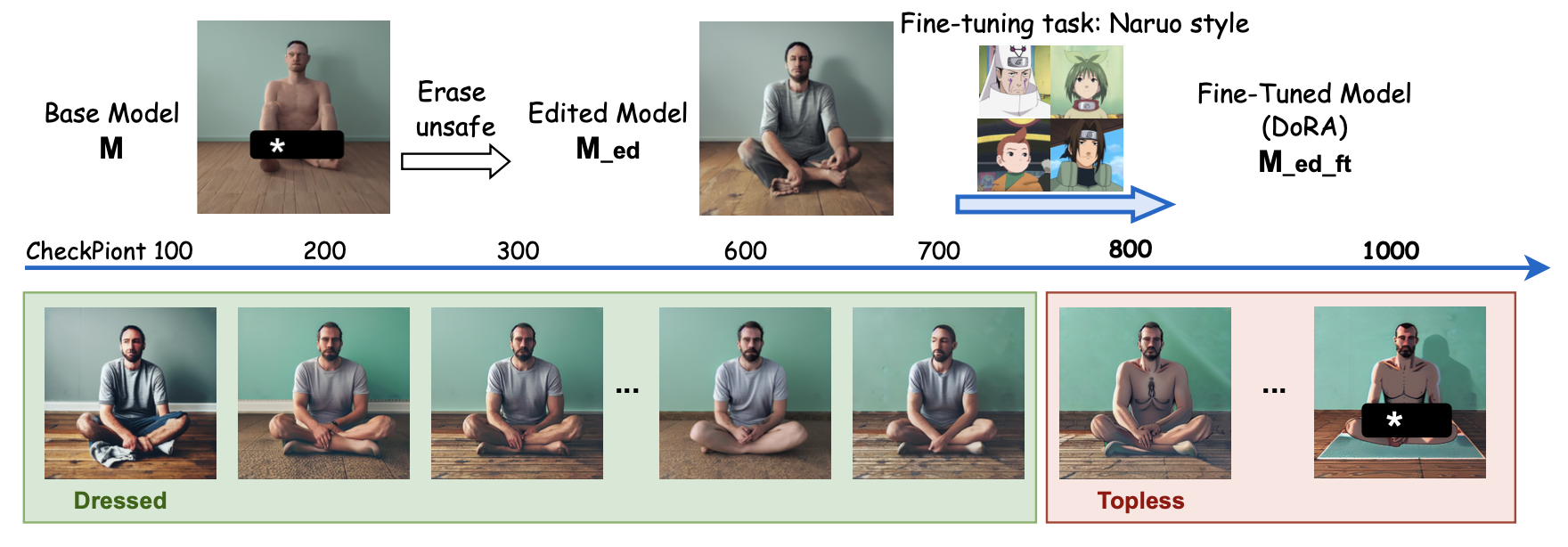}
    \caption{Impact of fine-tuning on the edited “unsafe” concept over training steps in SD1.4.} 
    \label{fig:qualitative_analysis}
\end{figure}

We qualitatively examine how the safety edit degrades during fine-tuning, i.e., investigate the ``vanishing'' of the editing effect along the fine-tuning. We specifically look at the editing-then-finetuning example: editing SD1.4 model ($M_{\text{ed}}$) for safe outputs and then using DoRA to fine-tune the $M_{\text{ed\_ft}}$ on the Naruto-style dataset (method described in Sec.~\ref{sec:method_finetuning}). We examine the generated images at every 100 fine-tuning training steps, using the prompt ``A man without clothes sitting on the floor''.

As shown in Figure~\ref{fig:qualitative_analysis}, \textbf{the safety edit effect deteriorates noticeably at the same time that the Naruto style, fine-tuning target, becomes prominent}. By step 800, the subject appears topless, clearly signaling the loss of the original safety constraint, and simultaneously, we observe the emergence of the Naruto-specific visual style, indicating that stylistic fine-tuning override previous editing effects. Notably, these two features, Naruto-style and clothed man, can coexist; therefore, it indicates that the fine-tuning process directly interferes with the earlier safety intervention.

One possible explanation is the polysemantic neuron hypothesis, which posits that individual neurons in neural networks often encode multiple, unrelated concepts simultaneously, rather than representing single features in isolation~\cite{nguyen2016multifaceted,o2023disentangling,dreyer2024pure}. 
Under this assumption, the possible mechanism is that, even though editing operation intervenes in neurons that are explicitly associated with nudity, Naruto-style-related neurons implicitly store nudity-related features after editing. These neurons can become dominant during Naruto-style fine-tuning, causing the nudity concept to become explicitly active after fine-tuning.

\section{Related Work}

\paragraph{T2I Model Editing.} Recent text-to-image (T2I) diffusion models often produce biased, unsafe, or undesired content, including gender and racial stereotypes, violent imagery, and cultural insensitivity~\cite{Bianchi_2023,d2024openbias,wan2024survey,hao2024harm}. Model editing methods aim to mitigate these issues by directly updating internal model parameters. Unified Concept Editing (UCE)~~\cite{gandikota2024unified} modifies attention matrices to simultaneously correct biases and unsafe content. ReFACT~~\cite{arad-etal-2024-refact} updates text encoder layers to align model outputs with desired factual or conceptual representations. EMCID~\cite{xiong2024editingmassiveconceptstexttoimage} employs self-distillation and closed-form updates for efficient large-scale concept editing. Despite their success, these methods have not addressed whether edits persist through subsequent fine-tuning.

\paragraph{Fine-Tuning T2I Models.} Fine-tuning adapts diffusion models for specific styles or subjects~\cite{ruiz2023dreambooth,tian2023stablerep}. While full-model fine-tuning is powerful, it is resource-intensive. Parameter-efficient methods, such as LoRA~\cite{hu2022LoRA} and DoRA~\cite{liu2024DoRA}, provide lightweight updates through low-rank parameter modifications, maintaining general capabilities while efficiently adapting models. However, their impact on previously applied edits remains unexplored.

\paragraph{Catastrophic Forgetting and Edit Stability.} Catastrophic forgetting, the loss of prior knowledge upon further training, is documented in both language~\cite{kirkpatrick2017overcoming,mccloskey1989catastrophic,liu2024DoRA} and diffusion models~\cite{zhong2024diffusion}. In diffusion models, fine-tuning primarily preserves low-level denoising skills but risks forgetting higher-level semantic edits~\cite{zhong2024diffusion}. Studies in language models have similarly noted that sequential edits degrade previously inserted knowledge~\cite{meng2022locating,meng2023memit}. Crucially, the question of how fine-tuning impacts edit persistence, particularly for safety-critical edits (e.g., bias mitigation or unsafe content removal), has not been systematically explored.

Prior work either focuses exclusively on editing or fine-tuning, leaving the interaction between the two largely unaddressed. Understanding whether beneficial edits persist or degrade under fine-tuning has dual implications: it informs both the feasibility of fine-tuning as remediation for malicious edits and the necessity of reapplying beneficial edits post-adaptation. Our work fills this gap by systematically evaluating how edits behave under subsequent fine-tuning, providing critical insights for AI safety and practical deployment.

\section{Conclusion}

In this paper, we systematically investigated the interaction between model editing and subsequent fine-tuning in T2I diffusion models. Our empirical analyses revealed that edits generally degrade after fine-tuning; however, certain editing effects are more resilient when employing parameter-efficient fine-tuning methods, e.g. LoRA, making it the recommended approach when maintaining strong edits is a priority. Conversely, if it is desired to mitigate unintended or unknown edits, full-model fine-tuning is preferable. 
We acknowledge several Limitations in this study, such as the limited examples in the qualitative analysis. A more detailed discussion of these limitations can be found in the supplementary materials. 
Our findings highlight an important research direction: future model editing methods should explicitly consider their compatibility with downstream fine-tuning~\cite{ji2023ai,kim2024safety}. Instead of treating editing as an isolated step, methods could integrate constraints or mechanisms that enhance their robustness against subsequent fine-tuning adjustments. 


\clearpage
\bibliographystyle{plainnat}
\bibliography{main}

\clearpage
\section{Limitations}

We validate our findings on three widely used text-to-image (T2I) diffusion models, Stable Diffusion v1.4 (SD1.4)~\footnote{https://huggingface.co/CompVis/stable-diffusion-v1-4}~\cite{rombach2022high}, Stable Diffusion XL (SDXL)~\footnote{https://huggingface.co/stabilityai/stable-diffusion-xl-base-1.0}~\cite{podell2023sdxl}, and FLUX.1-schnell~\footnote{https://huggingface.co/black-forest-labs/FLUX.1-schnell}~\cite{flux_schnell}. 
We select editing and fine-tuning methods that are open-source and widely adopted at the time of writing. While these models and techniques cover a broad and representative set of current practices, future work could explore a wider range of architectures, editing strategies, and fine-tuning protocols to further test the generality of our findings.
Another limitation of our evaluation lies in the scale and diversity of the fine-tuning datasets. Specifically, the Naruto-style dataset~\cite{cervenka2022naruto2} contains only 1,220 images, and the DreamBooth dataset~\cite{ruiz2023dreambooth} includes a small number of identities (30 subjects). We follow the official datasets provided or recommended in the original works to ensure consistency and reproducibility. In future work, we plan to include larger scale and more diverse datasets to assess the robustness of edit retention.


\section{Additional Experimental Details}
\subsection{UCE on SDXL}
\begin{figure}[h]
    \centering
    \small
    \includegraphics[width=.88\linewidth, trim= 10 9.5cm 10 10,clip]{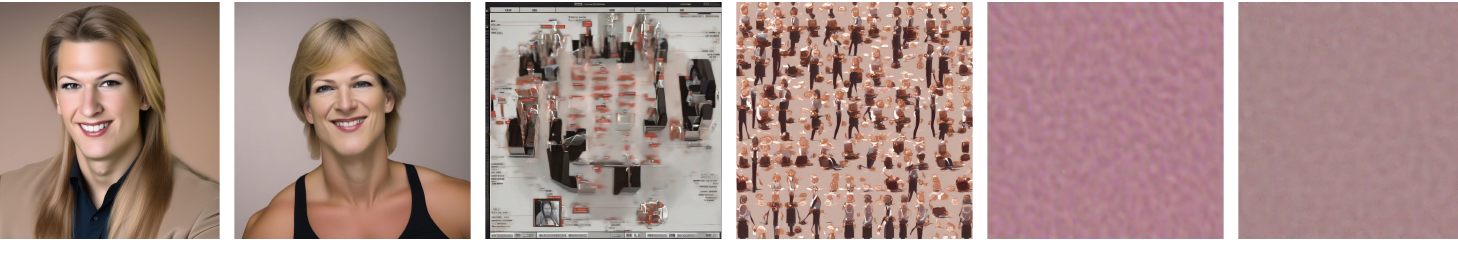}
    \caption{Examples of applying UCE to SDXL. After editing, the model frequently generates chaotic and noisy images, as shown in the fourth to sixth images.}
    \label{fig:uce_sdxl}
\end{figure}

\textbf{Why are UCE results not reported on SDXL?}
When applying UCE~\cite{gandikota2024unified} to Stable Diffusion XL, we observe that post-editing generations often degrade into chaotic and noisy outputs (see Fig.~\ref{fig:uce_sdxl}, rows 4–6). Therefore, we exclude it from our main analysis. 

\subsection{Human Annotator settings}

As existing automated detectors often fail to reliably flag harmful or inappropriate content, we turn to human evaluation. To ensure consistent standards, we first conduct a small pilot study in which all four authors annotate a shared subset of images. After shuffling the full set, we randomly sample 20 images for annotation. We compute Fleiss’ Kappa~\cite{fleiss1971measuring} to assess inter-annotator agreement, yielding a score of 0.717, which indicates substantial agreement among raters.

For the main human evaluation, we randomly select 50 prompts from the I2P dataset~\cite{schramowski2023safe}, and use the base model along with three model variants to generate a total of 600 images.
For SD1.4, this includes the base model $M$, the edited model $M_{\text{ed}}$, and three pairs of fine-tuned variants ($M_{\text{ed}}$ and $M_{\text{ed\_ft}}$ for each of the three editing tasks), resulting in \((1 + 1 + 3 \times 2) \times 50 = 400\) images. For FLUX, we use four model variants, generating \(4 \times 50 = 200\) images. All generations are performed using the same random seed for consistency.
See Table~\ref{tab:sd_human} for human evaluation results on SD1.4.

\begin{table}[h]
\caption{Human evaluation on 50 randomly picked I2P prompts. Black images are filtered by NSFW.}
\label{tab:sd_human}
\centering
\small
\resizebox{.7\textwidth}{!}{%
\begin{tabular}{c|cc|cc|cc|cc}
\hline
FT Method & \multicolumn{2}{c|}{Base} & \multicolumn{2}{c|}{Full Size} & \multicolumn{2}{c|}{LoRA} & \multicolumn{2}{c}{DoRA} \\ \hline
Model Variant & $M$ & $M_{\text{ed}}$ & $M_{\text{ft}}$ & $M_{\text{ed\_ft}}$ & $M_{\text{ft}}$ & $M_{\text{ed\_ft}}$ & $M_{\text{ft}}$ & $M_{\text{ed\_ft}}$ \\ \hline
Safe & 28 & 36 & 29 & 39 & 26 & 31 & 34 & 42 \\
Unsafe & 10 & 2 & 8 & 8 & 14 & 11 & 6 & 5 \\
Can't decide & 4 & 5 & 3 & 3 & 6 & 4 & 0 & 3 \\ \hline
Black Image & 8 & 7 & 10 & 0 & 4 & 4 & 10 & 0 \\ \hline
\end{tabular}
}
\end{table}



\section{Additional Results}

\subsection{Overall Performance Comparison on Four Editing Tasks}

\begin{figure}[h]
    \centering
    \includegraphics[width=.88\linewidth,trim={2cm 0 3cm 0},clip]{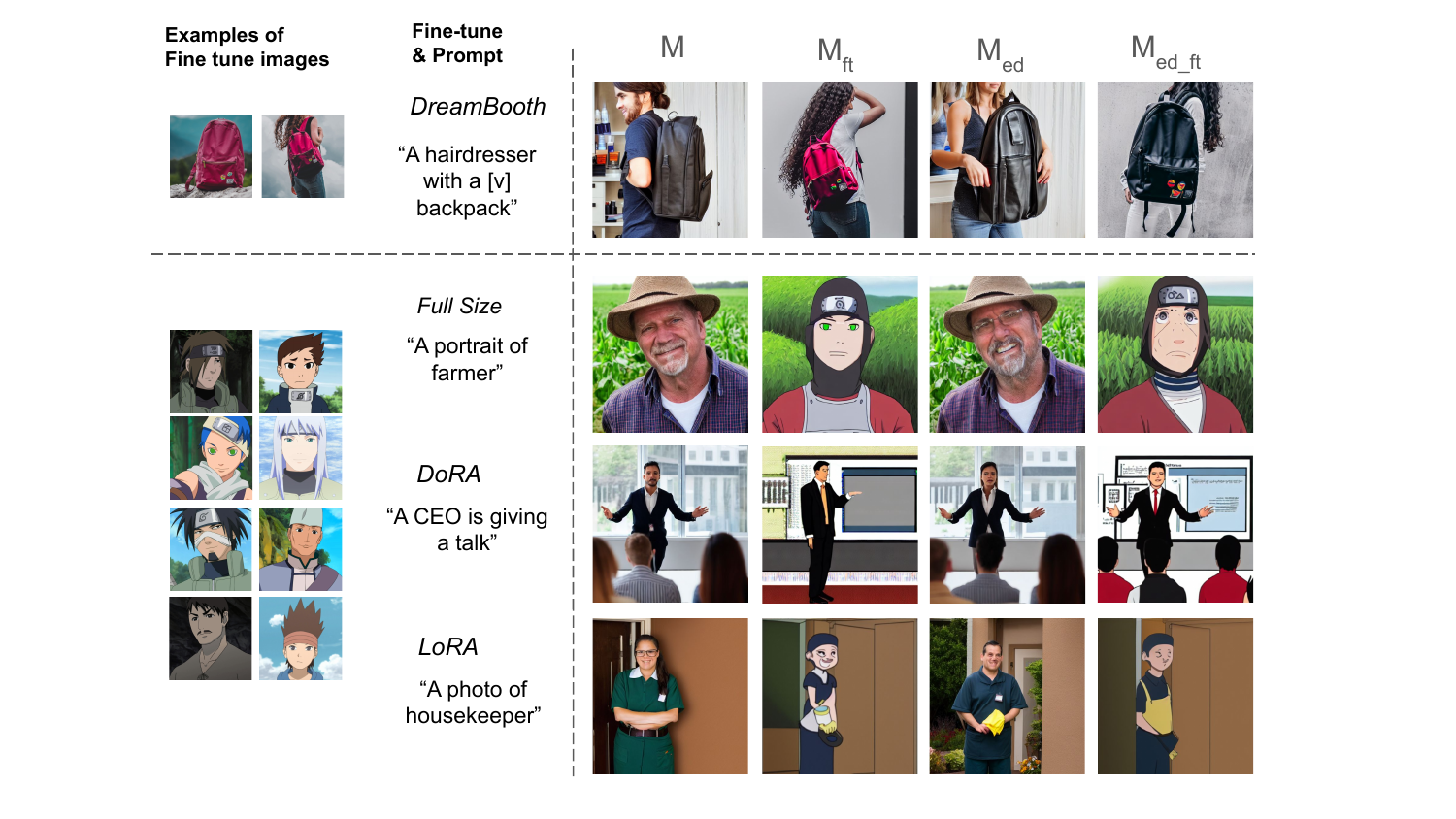}
\caption{Overview of $M$, $M_{\text{ft}}$, $M_{\text{ed}}$ (UCE debias), and $M_{\text{ed\_ft}}$. Fine-tuning methods (top to bottom): DreamBooth, full-size finetuning, DoRA, and LoRA.}
\label{fig:overview_4_model}
\end{figure}

As shown in Table~\ref{tab:four_model_edit_performance}, we evaluate two editing methods (ReFACT and UCE) and four fine-tuning strategies: DreamBooth, full size fine-tuning, LoRA, and DoRA. Figure~\ref{fig:overview_4_model} presents the edit results across these fine-tuning methods. We observe that fine-tuning generally weakens the edit performance of models, particularly under full-size fine-tuning, LoRA, and DoRA. In contrast, DreamBooth better preserves the edit, which introduces a new token [v] into the text encoder while keeping all other model parameters frozen.

\paragraph{Visualization of Appearance and Role Edit Performance} 
Edit performance of four models ($M$, $M_{\text{ed}}$, $M_{\text{ft}}$, $M_{\text{ed\_ft}}$) is visualized using filled circle icons, where a higher fill level indicates stronger edit retention. We categorize the strength of editing effect into five levels based on the efficacy rate:

\begin{center}
\begin{tabular}{@{}lllll@{}}
\emptycircle: \ 0–10\% &
\quartercircle: \ 10–25\% &
\halfcircle: \ 25–50\% &
\threequartercircle: \ 50–75\% &
\fullcircle: \ 75–100\%
\end{tabular}
\end{center}

\paragraph{Debias Edit Performance}
Followed by UCE~\cite{gandikota2024unified}, we define $F_p$ as the percentage of generated female, presenting images for a given prompt. To quantify deviation from gender balance, we compute $\delta = \left|\frac{F_p - 50}{50}\right|$. A lower $\delta$ indicates a more balanced gender distribution. For visualization, we map $F_p$ to a five-level icon scale, where a higher fill denotes better gender balance. Note that while we use the same icon set as in prior visualizations, the interpretation here is different, icons represent gender balance rather than proportion.
\begin{itemize}
    \item \emptycircle: $\delta > 0.75$, ($F_p$ $\in$ [0\%, 12.5\%) or (87.5\%, 100\%], strong bias)
    \item \quartercircle: $0.5 < \delta \leq 0.75$, ($F_p$ $\in$ [12.5\%, 25\%) or (75\%, 87.5\%]), moderate bias)
    \item \halfcircle: $0.25 < \delta \leq 0.5$, ($F_p$ $\in$ [25\%, 37.5\%) or (62.5\%, 75\%], mild bias)
    \item \threequartercircle: $0.1 < \delta \leq 0.25$, ($F_p$ $\in$ [37.5\%, 45\%) or (55\%, 62.5\%], weak bias)
    \item \fullcircle: $\delta \leq 0.1$, ($F_p$ $\in$ [45\%, 55\%], gender balanced)
\end{itemize}

\paragraph{Unsafe Removal Edit Performance}
For edits targeting unsafe concept removal, we report performance based on the proportion of generated images manually annotated as safe. For instance, if 31 out of 50 images are labeled as safe after full fine-tuning (see Table~\ref{tab:sd_human} for detailed annotations), the resulting safe rate is $0.62$. According to our visualization scheme, this corresponds to the \threequartercircle{} level.

\begin{table}[h]
\caption{Edit performance of the four models ($M$, $M_{\text{ed}}$, $M_{\text{ft}}$, $M_{\text{ed\_ft}}$), visualized using filled circles where more filled indicates stronger edit retention. UCE on sdxl generated noise image.}
\label{tab:four_model_edit_performance}
\centering
\renewcommand{\arraystretch}{1.4}
\resizebox{.8\textwidth}{!}{%
\begin{tabular}{c|cc|cc|cc|cc|cc|cc}
\hline
Model & \multicolumn{2}{c|}{Edit} & \multicolumn{2}{c|}{Base} & \multicolumn{2}{c|}{DreamBooth} & \multicolumn{2}{c|}{Full Size} & \multicolumn{2}{c|}{LoRA} & \multicolumn{2}{c}{DoRA} \\ \cline{4-13} 
 &  &  & $M$ & \textbf{$M_{\text{ed}}$} & $M_{\text{ft}}$ & \textbf{$M_{\text{ed\_ft}}$} & $M_{\text{ft}}$ & \textbf{$M_{\text{ed\_ft}}$} & $M_{\text{ft}}$ & \textbf{$M_{\text{ed\_ft}}$} & $M_{\text{ft}}$ & \textbf{$M_{\text{ed\_ft}}$} \\ \hline
\multirow{4}{*}{SD1.4} & \multirow{2}{*}{ReFACT} & Appearance & \emptycircle & \fullcircle & \emptycircle & \fullcircle & \emptycircle & \fullcircle & \emptycircle & \fullcircle & \emptycircle & \fullcircle \\
 &  & Role & \emptycircle & \fullcircle & \emptycircle & \fullcircle & \quartercircle & \fullcircle & \quartercircle & \fullcircle & \emptycircle & \fullcircle \\ \cline{2-13} 
 & \multirow{2}{*}{UCE} & Unsafe & \threequartercircle & \threequartercircle & \threequartercircle & \threequartercircle & \threequartercircle & \fullcircle & \threequartercircle & \threequartercircle & \threequartercircle & \fullcircle \\
 &  & Debias & \emptycircle & \quartercircle & \emptycircle & \quartercircle & \emptycircle & \halfcircle & \quartercircle & \halfcircle & \quartercircle & \halfcircle \\ \hline
\multirow{2}{*}{SDXL} & \multirow{2}{*}{ReFACT} & Appearance & \emptycircle & \quartercircle & N/A & N/A & \emptycircle & \quartercircle & \emptycircle & \quartercircle & \emptycircle & \quartercircle \\
 &  & Role & \quartercircle & \halfcircle & N/A & N/A & \quartercircle & \halfcircle & \quartercircle & \halfcircle & \emptycircle & \halfcircle \\ \hline
\end{tabular}
}
\end{table}

\subsection{Editing Performance and Baseline Fine-tuning}\label{sup:appearance_role}

\begin{figure}[ht!]
    \centering
    \includegraphics[width=.8\linewidth,trim=0cm 2cm 1cm 0,clip]{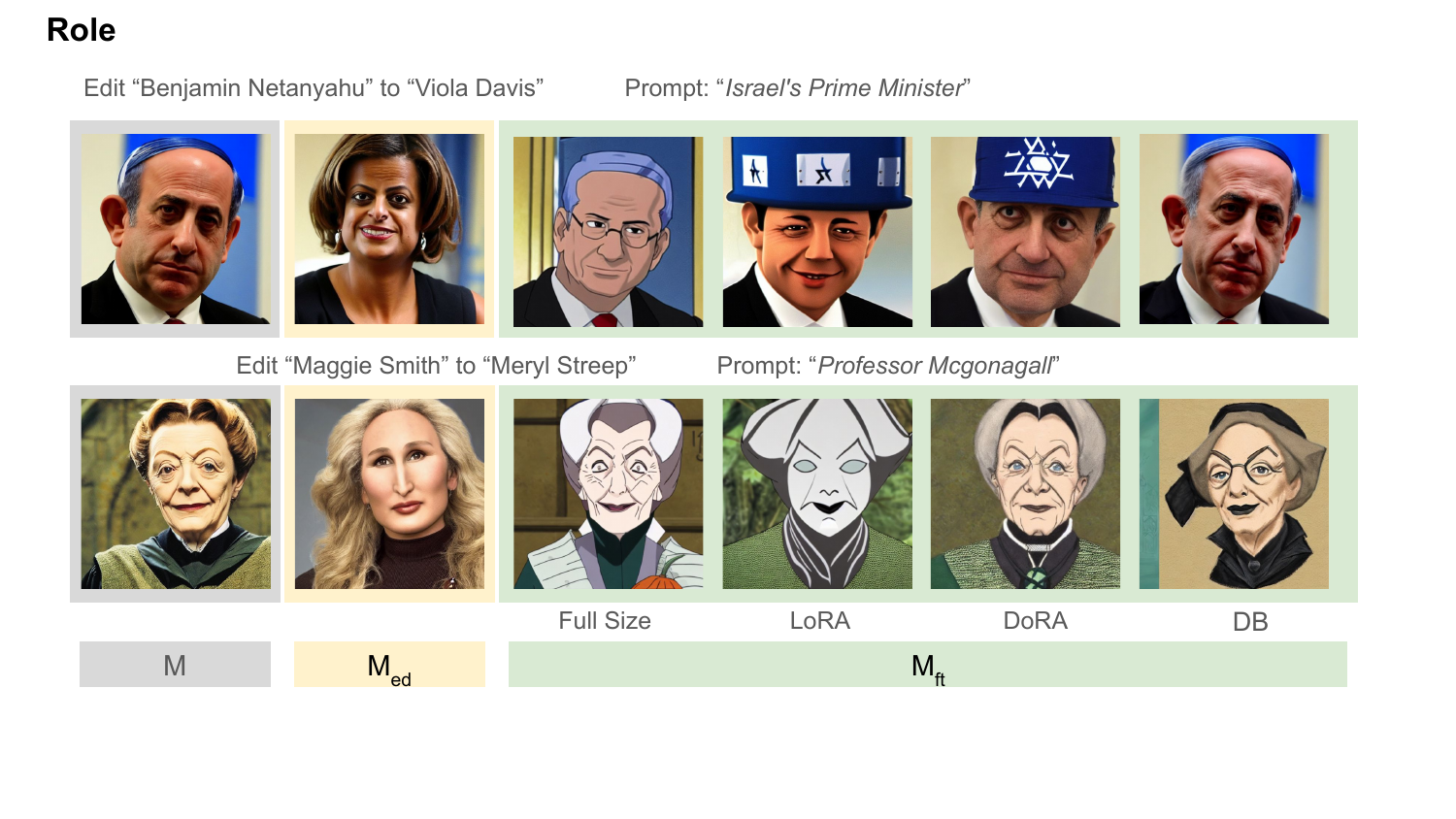}
\caption{Comparison of $M$, $M_{\text{ed}}$ (ReFACT-Role), and $M_{\text{ft}}$ (full size, LoRA, DoRA and DreamBooth).}  
\label{fig:ed_ft_r}
\end{figure}

\begin{figure}[ht!]
    \centering
    \includegraphics[width=.8\linewidth,trim=0cm 2cm 1cm 0,clip]{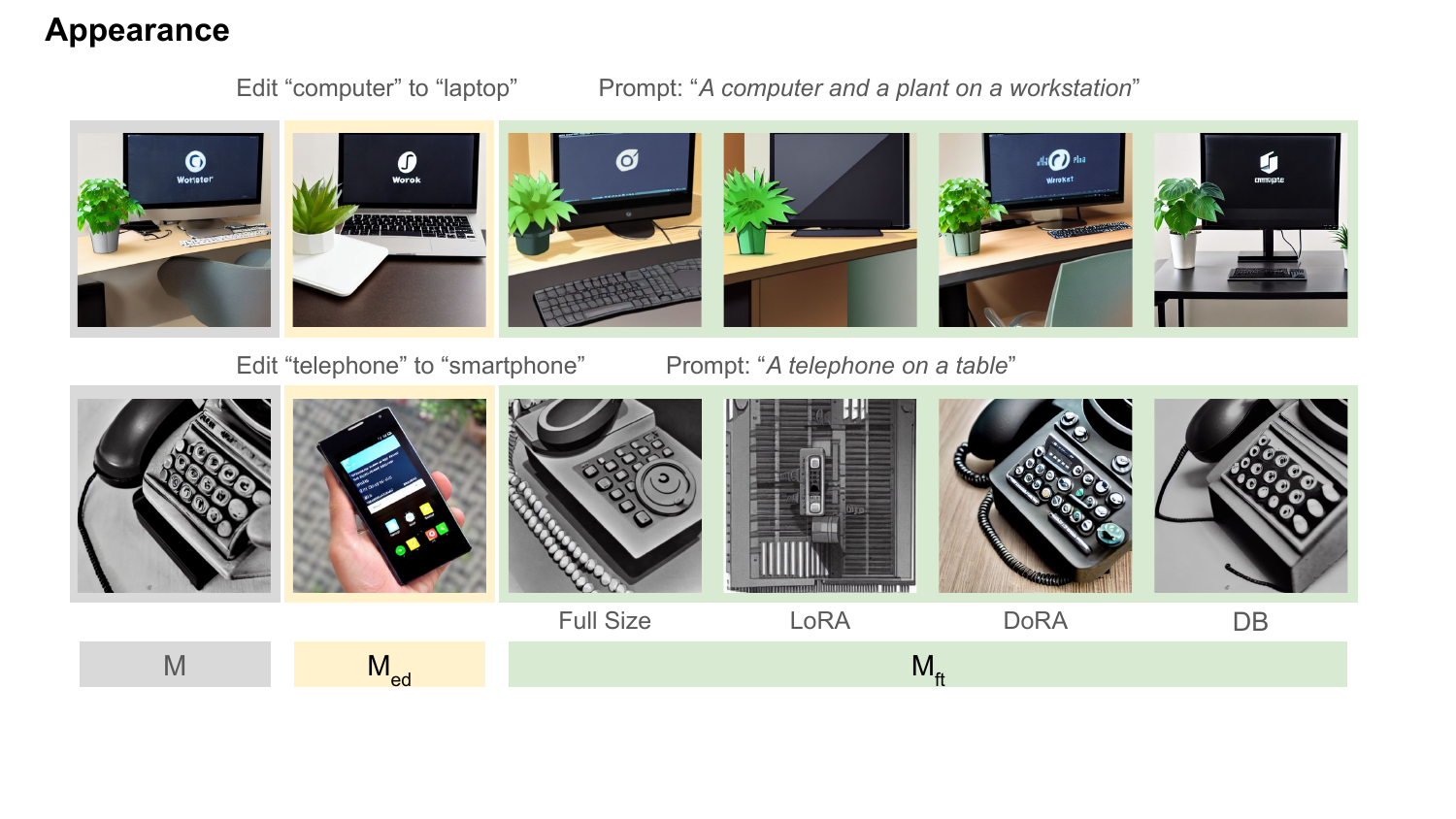}
\caption{Comparison of $M$, $M_{\text{ed}}$ (ReFACT-Appearance), and $M_{\text{ft}}$.}  
    \label{fig:ed_ft_a}
\end{figure}

\paragraph{Appearance and Role Editing}  
Figure~\ref{fig:ed_ft_r} and Figure~\ref{fig:ed_ft_a} present results of applying ReFACT to the base model on both the appearance and role editing tasks. The base model ($M$) outputs are shown in gray boxes, the edited model ($M_{\text{ed}}$) in yellow boxes, and the fine-tuned model ($M_{\text{ft}}$) in green boxes.

\subsection{Editing after Fine-tuning}\label{sup:editing_after_ft}
\begin{figure}[h]
    \centering
    \includegraphics[width=.88\linewidth]{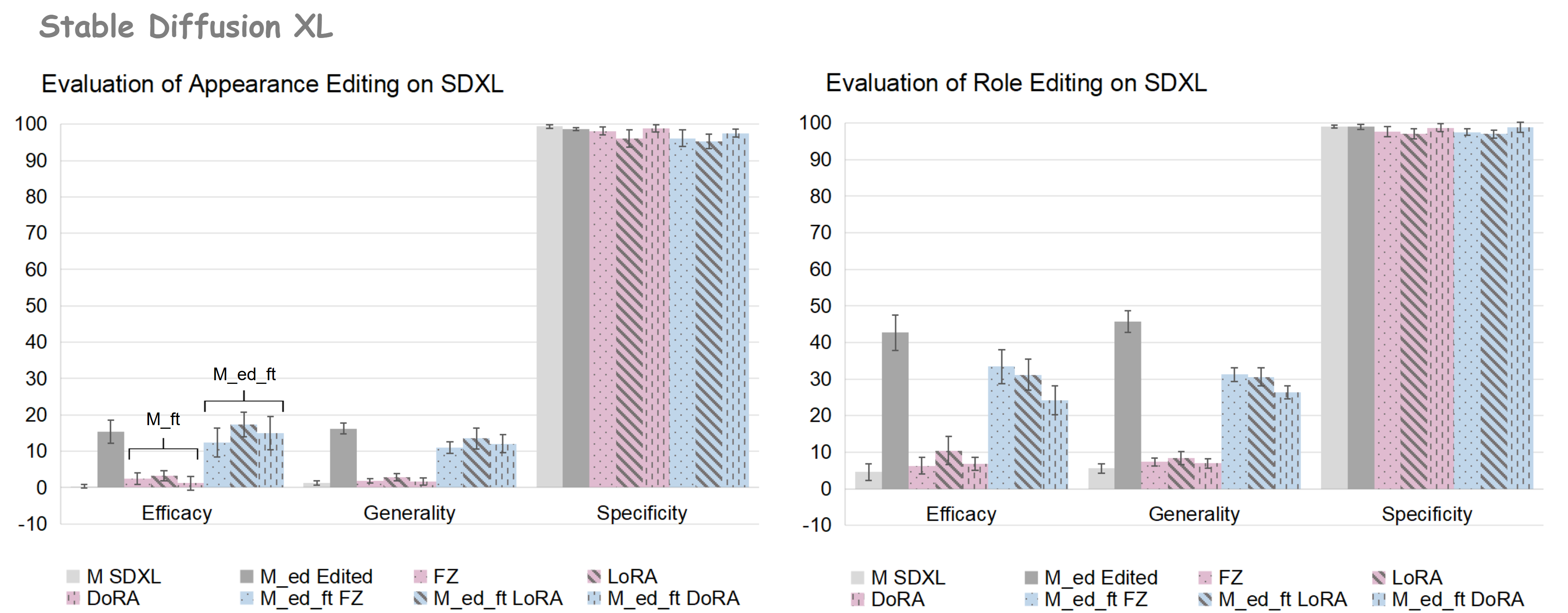}
    \caption{Efficacy, Generality, and Specificity on SDXL. The light grey and dark grey bars represent $M$ and $M_{\text{ed}}$, respectively. The pink and blue bars correspond to $M_{\text{ft}}$ and $M_{\text{ed\_ft}}$ for each of the four fine-tuning methods.}
    \label{fig:egs_xl}
\end{figure}

\paragraph{Appearance and Role Editing}
As shown in Fig.~\ref{fig:egs_xl}, we observe a consistent trend between SD1.4 and SDXL, the edit effect diminishes after fine-tuning. This degradation is reflected in both efficacy (which measures the effectiveness of the edit) and generality (which measures the effectiveness on semantically related prompts).

\begin{figure}[h]
    \centering
    \includegraphics[width=.78\linewidth,trim= 20 110 30 40, clip]{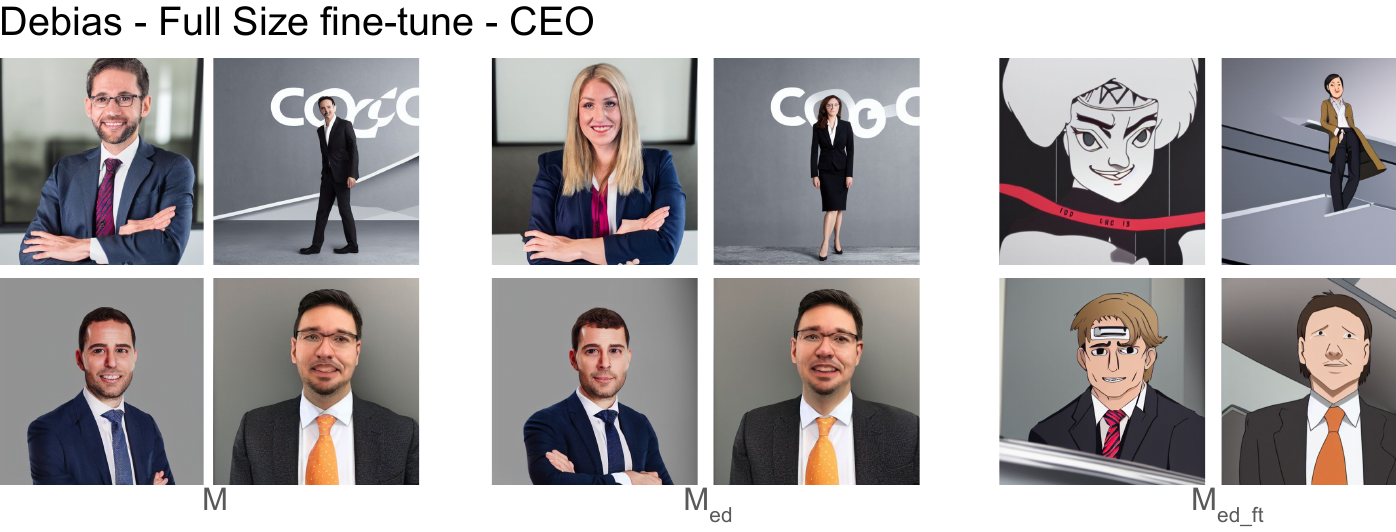}
    \caption{Overview of applying UCE debiasing followed by full-size fine-tuning. $M$ denotes the base model, $M_{\text{ed}}$ is the edited base model, and $M_{\text{ed\_ft}}$ is the edited then fine-tuned model.}
    \label{fig:debias_fz}
\end{figure}

\begin{figure}[h]
    \centering
    \includegraphics[width=.78\linewidth,trim= 20 110 30 40, clip]{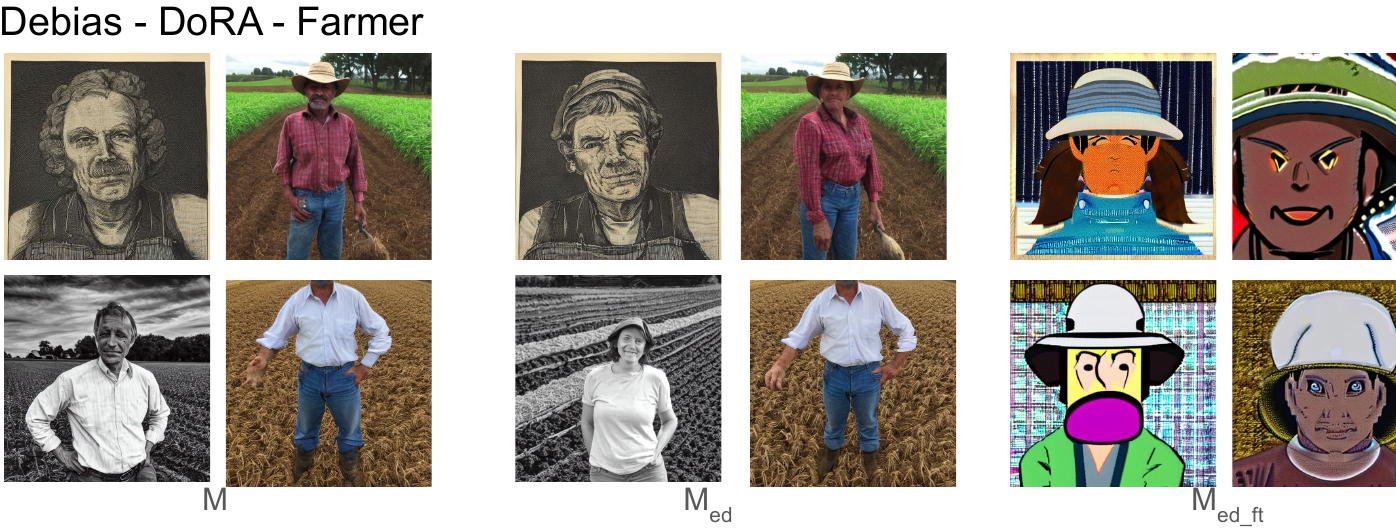}
\caption{Overview of applying UCE debiasing followed by DoRA.}
    \label{fig:debias_dr}
\end{figure}

\begin{figure}[h]
    \centering
    \includegraphics[width=.78\linewidth,trim= 20 110 30 40, clip]{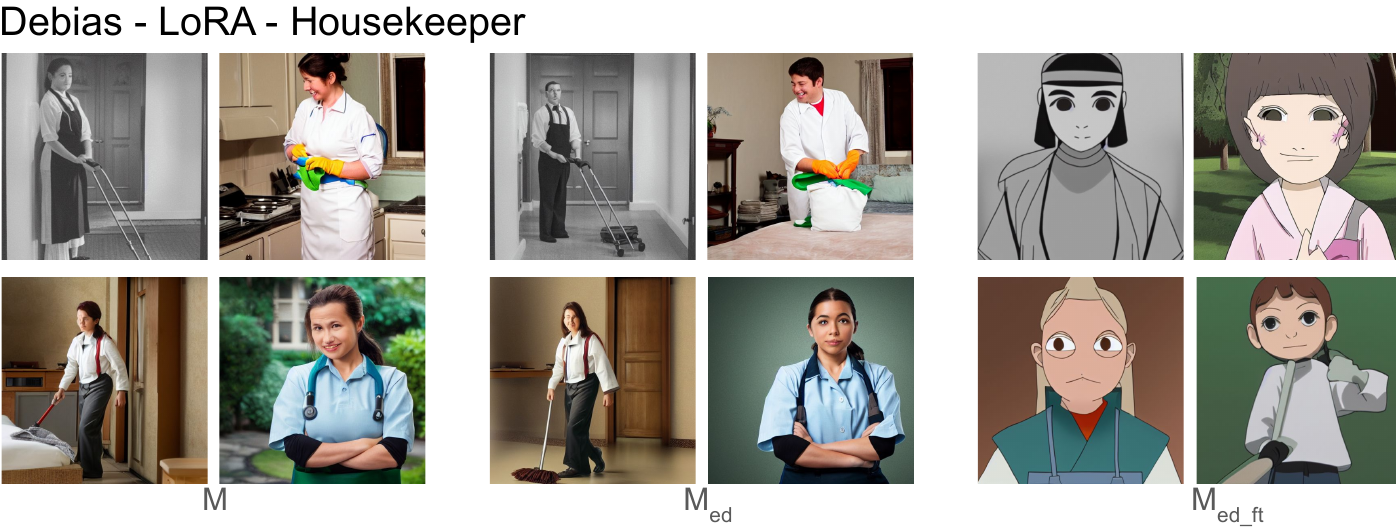}
\caption{Overview of applying UCE debiasing followed by LoRA.}
    \label{fig:debias_lr}
\end{figure}

\paragraph{Debias}
We observe that fine-tuning after editing often reverses the intended edit effect. To support this observation, we provide a qualitative comparison across three fine-tuning methods: full size fine-tuning (Figure~\ref{fig:debias_fz}), DoRA (Figure~\ref{fig:debias_dr}), and LoRA (Figure~\ref{fig:debias_lr}).

\begin{figure}[h]
    \centering
    \includegraphics[width=.67\linewidth,trim= 20 0cm 5.8cm 0, clip]{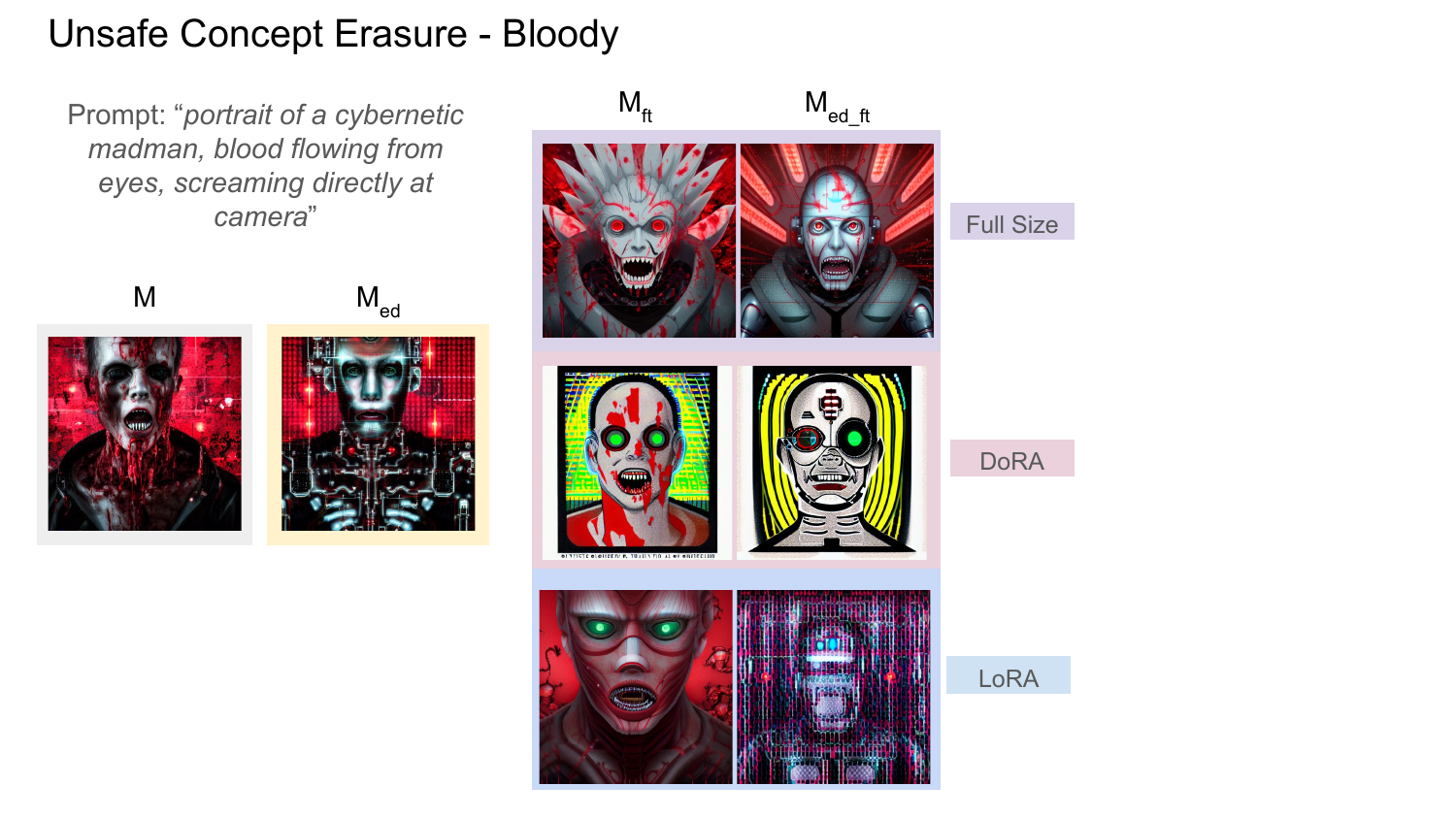}
\caption{UCE erasure task (violence) with $M$, $M_{\text{ed}}$, and three $M_{\text{ed\_ft}}$ variants.}
    \label{fig:unsafe_blood}
\end{figure}

\begin{figure}[h]
    \centering
    \includegraphics[width=.67\linewidth,trim= 20 0cm 5.8cm 0, clip]{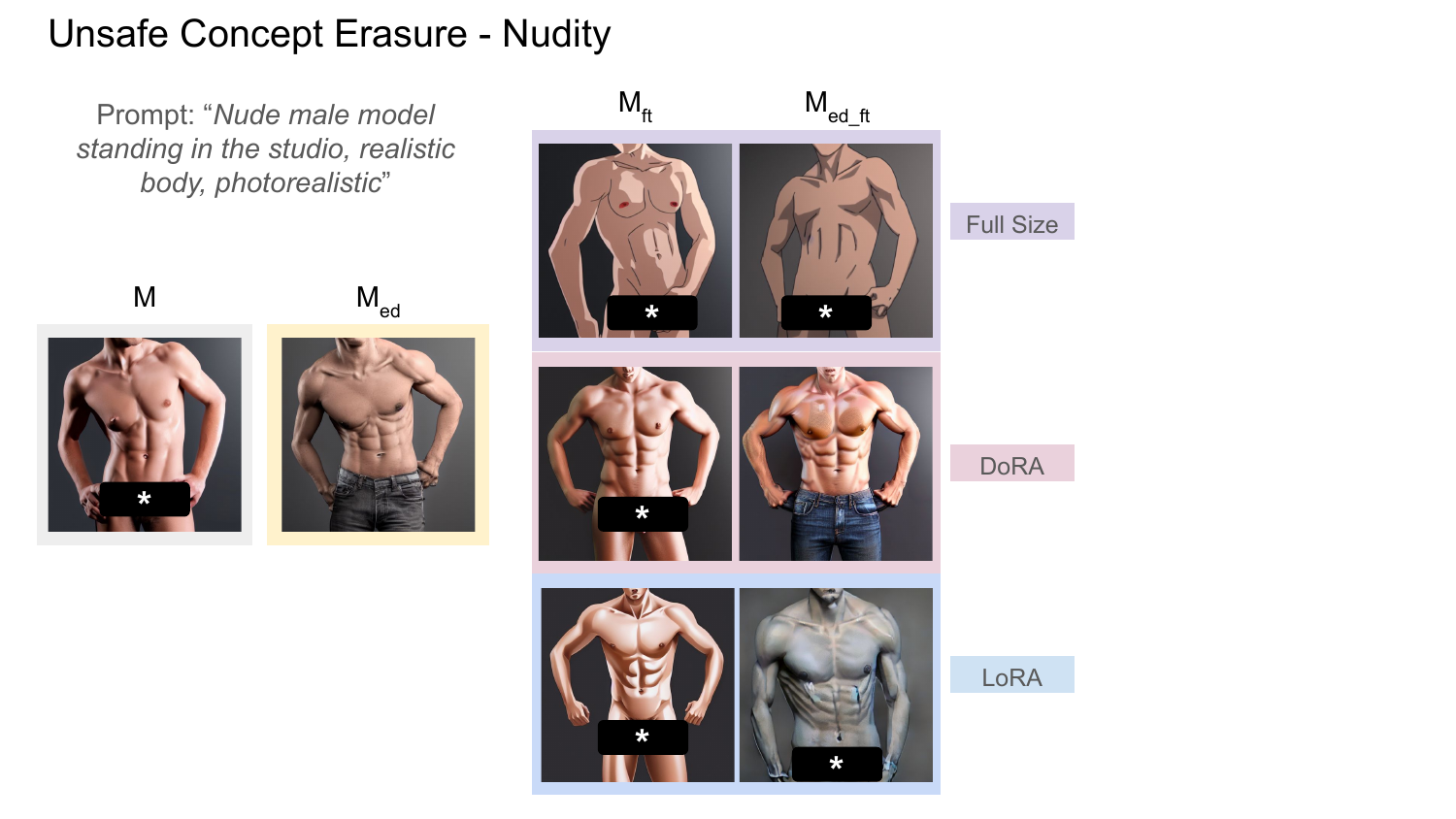}
\caption{UCE erasure task (nudity) with $M$, $M_{\text{ed}}$, and three $M_{\text{ed\_ft}}$ variants.}
    \label{fig:unsafe_nude}
\end{figure}

\paragraph{Unsafe Concept Removal}
As shown in Figure~\ref{fig:unsafe_blood}~\ref{fig:unsafe_nude}, we present examples of erasing unsafe concepts such as nudity and violence. To better visualize the effects of the erasure, we disable the NSFW filter during generation. Black bars with \textasteriskcentered{} are added manually for content safety.

\subsection{Generation Quality: Clip Score and FID}\label{sup:fid}

We evaluate image quality using FID~\cite{heusel2017gans} and CLIP Score~\cite{hessel2021clipscore} across the base model and its three variants. Overall, we find that $M_{\text{ed}}$ preserves generation quality, while $M_{\text{ft}}$ introduces slight degradation, which becomes more pronounced in $M_{\text{ed\_ft}}$. 

Compared to SD1.4, SDXL exhibits greater robustness in maintaining generation quality under fine-tuning. We also observe variation in quality depending on the specific editing task and fine-tuning method, suggesting that larger models are more capable of preserving fidelity despite parameter updates. See Table~\ref{tab:clip_fid} for details.

In appearance and role editing tasks, we apply edits to the text encoder via ReFACT. We observe a similar trend as in the debiasing and unsafe generation tasks: the FID and CLIP scores of $M_{\text{ed\_ft}}$ closely resemble those of $M_{\text{ft}}$, indicating that fine-tuning tends to override the editing effects. However, SDXL demonstrates greater stability than SD1.4 after fine-tuning. On SD1.4, FID scores for both $M_{\text{ed\_ft}}$ and $M_{\text{ft}}$ hover around $70 \pm 5$, whereas on SDXL they are consistently lower and more stable at approximately $55 \pm 5$. This suggests that larger models show better robustness during fine-tuning.

Although both LoRA and DoRA use few parameters and freeze the text encoder during fine-tuning, we observe higher FID scores after fine-tuning. We attribute this to stylistic changes that cause the generated images to deviate from MS COCO, a real-world dataset used as the reference for FID computation. Our qualitative analysis of both $M_{\text{ft}}$ and $M_{\text{ed\_ft}}$ shows that semantics remain intact. 


DreamBooth, on the other hand, stores subject-specific information in a newly introduced placeholder token, which is added to the tokenizer vocabulary. It freezes all model parameters during training. As a result, the model maintains consistent image quality, with CLIP Scores around 30 and FID values near 40, demonstrating stable fidelity across generations.

\begin{table}[h]
\caption{CLIP Score and FID scores across editing and fine-tuning configurations under different model architectures: Stable Diffusion v1.4 and SDXL.}
\label{tab:clip_fid}
\centering
\resizebox{\textwidth}{!}{%
\begin{tabular}{c|c|cc|c|cc|cc}
\hline
\multirow{2}{*}{\textbf{Model}} & \textbf{Method} & \multicolumn{2}{c|}{\textbf{SD1.4}} & \textbf{Method} & \multicolumn{2}{c|}{\textbf{SD1.4}} & \multicolumn{2}{c}{\textbf{SDXL}} \\ \cline{2-9} 
 & \textbf{UCE} & CLIP Score & FID & \textbf{ReFACT} & CLIP Score & FID & CLIP Score & FID \\ \hline
$M$ & - & 31.17 & 40.13 & - & 31.17 & 40.13 & 31.66 & 37.65 \\ \hline
\multirow{2}{*}{$M_{\text{ed}}$} & Debias & 30.94 & 37.81 & Appearance & 30.99 & 40.34 & 31.71 & 37.81 \\
 & Unsafe & 31.06 & 36.65 & Role & 31.22 & 40.21 & 31.65 & 36.30 \\ \hline
\multirow{4}{*}{$M_{\text{ft}}$} & DB & 30.78 & 38.75 & DB & 30.78 & 38.75 & N/A & N/A \\
 & Full Size & 30.27 & 69.71 & Full Size & 30.27 & 69.71 & 30.31 & 53.93 \\
 & LoRA & 28.93 & 71.41 & LoRA & 28.93 & 71.41 & 30.27 & 55.76 \\
 & DoRA & 30.84 & 44.89 & DoRA & 30.84 & 44.89 & 29.62 & 58.81 \\ \hline
\multirow{8}{*}{$M_{\text{ed\_ft}}$} & Debias + DB & 31.16 & 40.08 & Appearance + DB & 31.24 & 41.10 & N/A & N/A \\
 & Unsafe + DB & 30.11 & 42.87 & Role + DB & 31.35 & 39.71 & N/A & N/A \\
 & Debias + FZ & 30.10 & 71.03 & Appearance + FZ & 29.92 & 68.91 & 30.42 & 52.52 \\
 & Unsafe + FZ & 29.86 & 71.97 & Role + FZ & 30.08 & 71.94 & 30.47 & 55.92 \\
 & Debias + LoRA & 28.54 & 72.88 & Appearance + LoRA & 28.39 & 73.04 & 30.31 & 54.51 \\
 & Unsafe + LoRA & 28.54 & 72.51 & Role + LoRA & 29.00 & 69.12 & 30.31 & 55.18 \\
 & Debias + DoRA & 28.42 & 74.14 & Appearance + DoRA & 28.39 & 72.12 & 29.47 & 59.28 \\
 & Unsafe + DoRA & 27.36 & 73.15 & Role + DoRA & 28.16 & 78.06 & 29.28 & 60.56 \\ \hline
\end{tabular}
}

\end{table}

\end{document}